\documentclass{article}

     \PassOptionsToPackage{numbers, compress}{natbib}

     \usepackage[preprint]{neurips_2019}

\usepackage{multirow}
\usepackage[utf8]{inputenc} 
\usepackage[T1]{fontenc}    
\usepackage{hyperref}       
\usepackage{url}            
\usepackage{lipsum}
\usepackage{amsmath}
\usepackage{booktabs}       
\usepackage{amsfonts}       
\usepackage{nicefrac}       
\usepackage{microtype}      
\usepackage{graphicx}
\usepackage{natbib}
\usepackage{color}
\usepackage{float}
\usepackage{array}
\usepackage{booktabs}
\usepackage{authblk}
\usepackage[misc]{ifsym}
\setcitestyle{square,comma}
\newcommand{\etal}{\textit{et al}.}
 \normalsize

\renewcommand{\arraystretch}{2}

\newcommand{\zttian}[1]{{{#1}}}

\newcommand{\ma}[1]{\mathrm{#1}} 

\title{Region Refinement Network for Salient Object Detection}

\begin{document}

\author{
\textbf{
Zhuotao Tian\textsuperscript{\,1} \enspace 
Hengshuang Zhao\textsuperscript{\,1} \enspace 
Michelle Shu\textsuperscript{\,2} \enspace 
Jiaze Wang\textsuperscript{\,1} \\}
\vspace{-0.3cm}
\textbf{
Ruiyu Li\textsuperscript{\,3} \enspace 
Xiaoyong Shen\textsuperscript{\,3}\enspace
Jiaya Jia\textsuperscript{\,1,3}\\
}
\vspace{-0.2cm}
\textsuperscript{\,1}The Chinese University of Hong Kong \enspace 
\textsuperscript{\,2}John Hopkins University\\
\vspace{-0.3cm}
\textsuperscript{\,3}Youtu Lab, Tencent \\
\vspace{0.1cm}
{\tt\small \{zttian,hszhao,jzwang,leojia\}@cse.cuhk.edu.hk \enspace 
\tt\small shu.michelle97@gmail.com \quad
\tt\small \{royryli,dylanshen\}@tencent.com }
}

\maketitle
\vspace{-0.7cm}
\begin{abstract}
Albeit intensively studied, false prediction and unclear boundaries are still major issues of salient object detection. In this paper, we propose a Region Refinement Network (RRN), which recurrently filters redundant information and explicitly models boundary information for saliency detection.
Different from existing refinement methods, we propose a Region Refinement Module (RRM) that optimizes salient region prediction by incorporating supervised attention masks in the intermediate refinement stages. The module only brings a minor increase in model size and yet significantly reduces false predictions from the background. To further refine boundary areas, we propose a Boundary Refinement Loss (BRL) that adds extra supervision for better distinguishing foreground from background. 
BRL is parameter free and easy to train. We further observe that BRL helps retain the integrity in prediction by refining the boundary.
Extensive experiments on saliency detection datasets show that our refinement module and loss bring significant improvement to the baseline and can be easily applied to different frameworks.
We also demonstrate that our proposed model generalizes well to portrait segmentation and shadow detection tasks.

\end{abstract}

\vspace{-0.5cm}
\section{Introduction}
Salient objects are one primary information source in the human perception of natural scenes. Salient Object Detection (SOD), as an essential part of computer vision, aims at identifying the salient objects in wild scenes. Previous SOD methods, as fundamental tools, have benefited a wide range of applications. Topics that are closely or remotely related to visual
saliency include semantic segmentation \cite{zhao2017pspnet,deeplab,deeplabv3,rr}, instance segmentation \cite{maskrcnn,panet,masklab}, object detection\cite{faster-rcnn,intro_detection1,intro_detection2,intro_detection3}, to name a few. 

Early SOD methods utilize low-level, hand-crafted features that are vulnerable to complex scenarios. Recently, deep learning based methods \cite{exp1,exp2,exp3,exp4} greatly improved performance on benchmark datasets. However, this task is still challenging in natural scenes due to the varying attributes such as tiny and thin structures, various background and multiple objects. 

To alleviate these difficulties, the work of \cite{r3net,srm} used recurrent structures to refine results. These methods do not control the information flow during the recurrent refinement process and allow redundant information pass, possibly resulting in performance reduction, which will be discussed more in Section \ref{sec:effect_grm}. To selectively pass useful information for better results, we design a cascaded region refinement module with a gating mechanism that can be viewed as an attention map. It is guided by the ground truth to control information flow and helps the network focus on refining only salient regions. Our extensive experiments in Section \ref{sec:otherframeworks} show that this design is general for applying to other frameworks and improves their respective performance.

Common boundary area errors on salient object predictions are also addressed. As shown later in Figure \ref{fig:error_map}(d)\&(f), absolute errors between ground truth and predicted saliency maps mainly stem from the boundary area. To obtain a better prediction, we propose Boundary Refinement Loss (BRL) to help refine boundary areas. Compared with previous boundary strategies \cite{exp6,contour}, BRL is more efficient because it does not require additional parameters and computation. Although the method of \cite{nonlocalsalient} also considers boundary enhancement, our ablation study in Section \ref{sec:effect_BRL} show that our strategy achieves better performance because both foreground and background near the boundary are supervised and optimized.

Our contribution in this paper is threefold.
\vspace{-0.1in}
\begin{itemize}
    \item We propose a new effective region refinement module.
    \item We propose an efficient enhancement strategy to refine prediction on the boundary.
    \item Our model achieves new state-of-the-art results on five representative benchmark datasets. Also, our designs can be easily applied to other frameworks.
\end{itemize}

\vspace{-0.2cm}
\section{Related Work}
\vspace{-0.1cm}
\label{relatedworks}
In recent years, salient object detection has attracted much attention due its wide range of applications \cite{exp1,exp2,salientinstance,exp3,reslt,exp4,wang2018ProgAtten,exp6,exp7,DUTS,exp9,exp10,exp11,poolnetcvpr2019,exp12,exp13,tian2022gfsseg,exp14,exp15,lai2021cac,exp16,exp17,lai2022decouplenet,bottomupsalient,jiang2021semi,exp18, cuhk,automata, intersalient, intellmedia,tian2020pfenet,Tian_2019_CVPR}. Methods designed for salient object detection can be divided into two categories of deep learning based methods and those without learning. Non-deep-learning methods are primarily based on low-level features like regional contrast \cite{ECSSD,contrast1,contrast2,contrast3,contrast4} and priors \cite{prior1,OMRON,prior3,prior4,prior5}. More details can be found in \cite{survey}. In this section, we mainly discuss deep learning based solutions.

Early deep learning methods use features generated by deep neural networks (DNNs) to calculate saliency of image units, like super-pixels and proposals. For example, in \cite{HKU-IS}, with the features extracted from DNN, the score of each super-pixel of an image can be easily calculated. In \cite{superpixel2}, two networks are utilized (one for local superpixel estimation and the other for global proposal search). The final saliency map is generated by a weighted sum of salient object regions. 

Later, end-to-end frameworks \cite{amulet,DCL,DSS,DSS2,r3net,recurAtten,wang2018BiDirectional,liu2018PicaNet,srm,DHSNet} became the mainstream and many effective structures were proposed. In \cite{amulet}, multi-level features are aggregated from a backbone network for salient object detection. \cite{DCL} proposes a deep contrast network with a pixel-level fully convolutional stream and a segment-wise spatial pooling module to deal with blurry regions. \cite{DSS} introduces short connections to the skip-layer structures within the holistically-nested edge detector (HED) \cite{HED} to boost performance. \cite{wang2018BiDirectional} designs a bi-directional structure that enables information flow between lower and higher layers, and \cite{liu2018PicaNet} proposes an attention mechanism that helps capture contextual information for each pixel in both local and global forms. \cite{DHSNet} proposes to hierarchically and progressively refine the saliency map with local context information. Our work is different in that we use recurrent structures and exploit boundary enhancement.

Recurrent structures have been proved useful \cite{recurAtten,r3net,srm}. In R3Net \cite{r3net}, Residue Refinement Blocks (RRBs) are recurrently applied to the extracted deep feature to refine the output. \cite{recurAtten} combines spatial transformer and recurrent network units to iteratively perform saliency refinement on sub-regions of images. \cite{srm} leverages a stage-wise refinement network to refine saliency prediction in a coarse-to-fine manner. Our method is inherently different from the above methods. With a novel supervised mask, our proposed cascaded refinement module involves the attention-like mechanism. It significantly improves performance by keeping meaningful information. 

On the other hand, boundary (or contour) is essential to consider \cite{contour,exp6,nonlocalsalient,boundary_aware}. C2S-Net \cite{contour} has two branches: one predicts contour and the other generates saliency maps with a contour-to-saliency transferring method to utilize contour learning in saliency map prediction. \cite{exp6} proposes a local Boundary Refinement Network to adaptively learn the local contextual information for each pixel, which helps capture relationships in neighbors. \cite{boundary_aware} uses a separate branch to generate boundary prediction and fuses it with a region saliency map to refine results. Both methods involve additional parameters to form a new branch for boundary prediction. Differently, \cite{nonlocalsalient} proposes an additional IoU Boundary Loss generated by gradients of ground truth and predicted salient region map. Likewise, our boundary refinement strategy only involves an additional loss, but it directly optimizes the boundary region rather than gradients, resulting in even better performance.

\vspace{-0.2cm}
\section{Our Method}
\vspace{-0.2cm}
In this section, we motivate our solution and detail the new Region Refinement Module (RRM) and Boundary Refinement Loss (BRL), vastly useful in saliency detection. Then we propose the Region Refinement Network (RRN).

\vspace{-0.2cm}
\subsection{Region Refinement Module}
\label{sec:GRM}
\vspace{-0.1cm}
Recently proposed refinement modules refine input features without any selection process. We note input features always bring redundant information that may cause error or ambiguity. For example, for features extracted from multiple layers \cite{DSS,cheng2019DNA,r3net}, pixels belonging to tiny structures, such as animal hair and tail, need additional spatial information, while pixels for large objects like elephant body and airplane need more semantic clues. Without carefully handling this issue, the final prediction may contain hard samples for false positives with high probability and false negatives with low probability. To alleviate this difficulty, we propose Region Refinement Module (RRM) in the cascaded structure that helps accumulate useful information to the final feature. In addition, different from previous work in Section \ref{relatedworks}, we utilize the output salient map in each intermediate refinement stage as an attention map to guide the information flow. Our experiments with/without using the mask in Section \ref{sec:effect_grm} show that this gating mechanism can refine features in a particular way for better overall performance.

\begin{figure}
\centering
    \begin{minipage} [t] {0.7\linewidth}
        \centering
        \includegraphics [width=1\linewidth,height=0.55\linewidth] {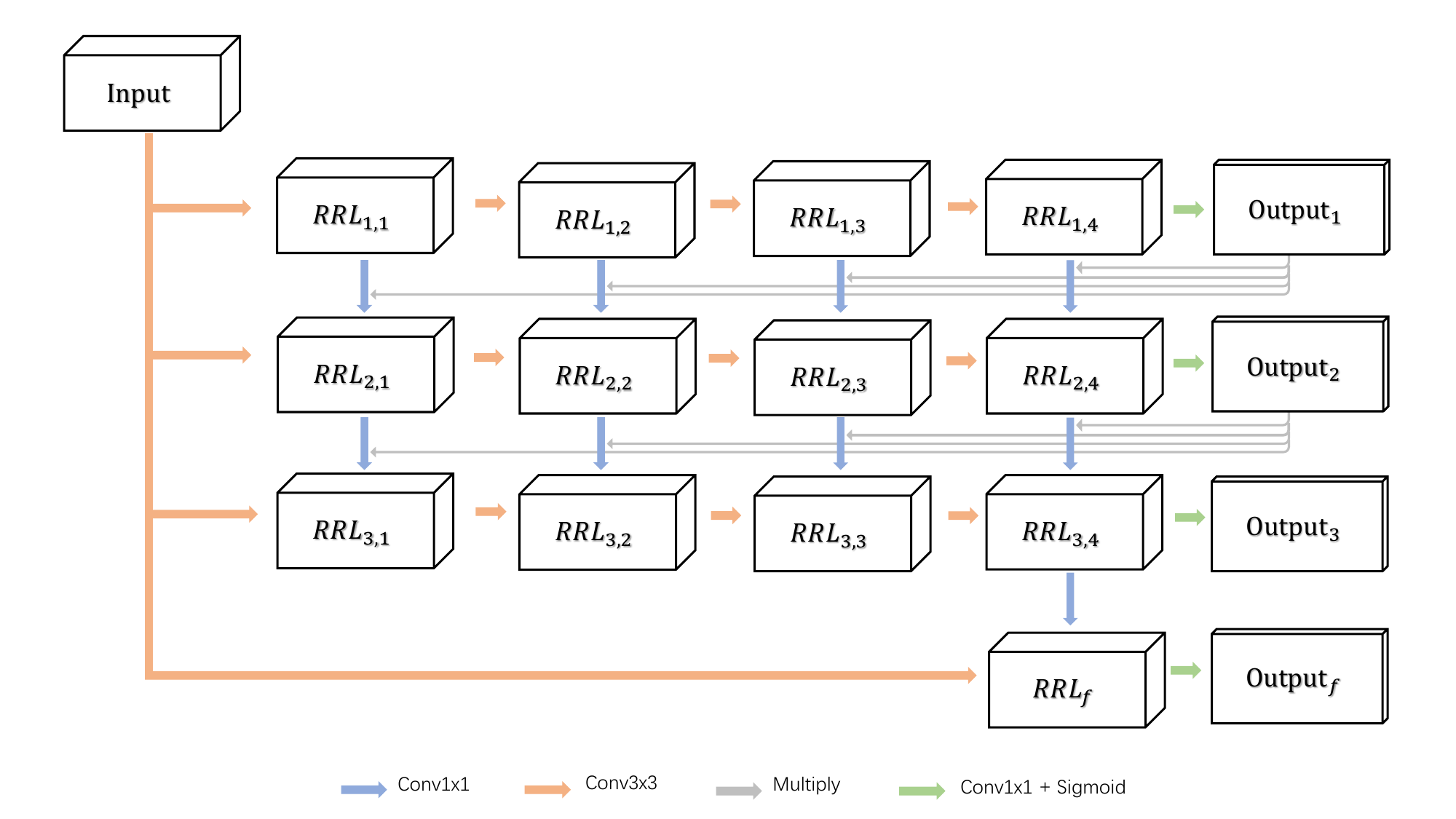}
    \end{minipage}
    \vspace{-0.2cm}
    \caption{Example of Region Refinement Module ($N=3, M=4$).}
    \label{fig:GRM}
     \vspace{-0.5cm}
\end{figure}

\vspace{-0.2cm}
\subsubsection{Module Structure}
\vspace{-0.1cm}
As shown in Figure \ref{fig:GRM}, RRM is composed of several Region Refinement Layers (RRL). $N$ is the number of RRL, and $M$ is the number of 3$\times$3 convolutions (equals to the number of 1$\times$1 convolutions) in each RRL.
\zttian{Each RRL receives coarse features from the input and previous layers and passes the refined features to the next layer. At the end of each RRL, the output is processed by a Sigmoid function. It is then used as an attention mask to gate the refined feature to the next layer, making the model focus more on refining salient pixels.}

\begin{equation}
\label{eqn:grl_ori}
RRL_{i,j}=\left\{
\begin{array}{lcl}
\mathcal{G}(\ma{X}) & & {i=1,j=1}\\
\mathcal{G}(\mathcal{F}(RRL_{i-1,j} \odot \ma{Y}_{i-1}) + \ma{X}) & & {i>1,j=1}\\
\mathcal{G}(RRL_{i,j-1}) & & {i=1,j>1}\\
\mathcal{G}(\mathcal{F}(RRL_{i-1,j} \odot \ma{Y}_{i-1}) + RRL_{i,j-1}) & & \textrm{Otherwise}
\end{array} \right.
\end{equation}

\begin{equation}
\label{eqn:output}
\ma{Y}_{i} = \mathcal{F_{\rho}}(RRL_{i,M})  \hspace{5mm}  i=1,2,......,N
\end{equation}

The formulation of RRM is shown in Eqs. \eqref{eqn:grl_ori} and \eqref{eqn:output} where $\ma{X}$ is the input feature,  $\ma{Y}_{i}$ is the output from $RRL_i$, $N$ is the number of layers, $M$ is the number of 3$\times$3 convolutions, $\odot$ denotes pixel-wise multiplication, $\mathcal{F}$ represents $1\times1$ convolution with ReLU function, $\mathcal{G}$ represents $3\times3$ convolution with ReLU function and $\mathcal{F}_{\rho}$ represents $1\times1$ convolution with Sigmoid function. 

As Eq. \eqref{eqn:grl_ori} shows, output $\ma{Y}_{i}$ from current layer $RRL_i$ is used as a gate to control the information passed to next layer $RRL_{i+1}$ by a pixel-wise multiplication, which helps filter out redundant feature with low probability and retain the information of interest with high probability in the previous stage.

\begin{equation}
\label{eqn:grl_f}
RRL_f = \mathcal{F}(U(RRL_{N,M})) +  \mathcal{G}(U(\ma{X}))
\end{equation}
\begin{equation}
\label{eqn:final_output}
\ma{Y}_f = \mathcal{F_{\rho}}(RRL_f)
\end{equation}
At last layer $RRL_N$, as Eqs. \eqref{eqn:grl_f} and \eqref{eqn:final_output} show ($U$ is the upsampling function), $RRM$ generates final output $\ma{Y}_{f}$ by upsampling $RRL_{N,M}$ and input feature twice larger then applying 1x1 convolution to both.

In $RRM$ we calculate cross entropy loss for all outputs. Since $RRM$ only takes $\ma{Y}_{f}$ as the final output, the final loss combines $\ma{Y}_{f}$  and $\ma{Y}_{i}$  as
\begin{equation}
\label{eqn:loss}
L = L_f + \frac{\lambda}{N}\sum_{i=1}^{N}L_i = L_f + \sigma\sum_{i=1}^{N}L_i
\end{equation}
where $L_f$ is the cross entropy loss of $\ma{Y}_{f}$, $L_i$ is the cross entropy loss of $\ma{Y}_{i}$. $\sigma$ is used for balancing the loss of previous outputs. $\sigma$ equals $\frac{\lambda}{N}$ where $N$ is the number of $RRL$ and $\lambda$ is a balancing factor for $\ma{Y}_{i}$. The effect of $\sigma$ is discussed in Section \ref{sec:effect_grm}.

\vspace{-0.4cm}
\subsection{Boundary Refinement Loss}
\vspace{-0.2cm}
\subsubsection{Motivation}
\vspace{-0.1cm}

\begin{figure}
    \begin{minipage} [t] {0.12\linewidth}
        \centering
        \includegraphics [width=1\linewidth,height=0.75\linewidth] 
        {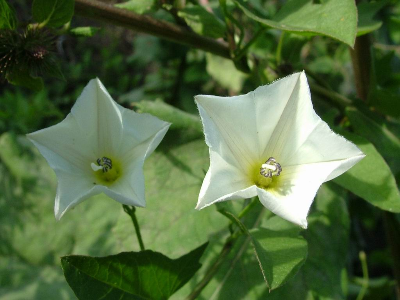}
    \end{minipage}
    \begin{minipage} [t] {0.12\linewidth}
        \centering
        \includegraphics [width=1\linewidth,height=0.75\linewidth] {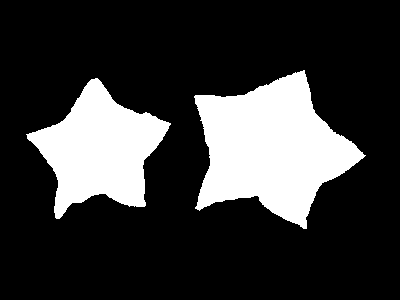}
    \end{minipage}
    \begin{minipage} [t] {0.12\linewidth}
        \centering
        \includegraphics [width=1\linewidth,height=0.75\linewidth] {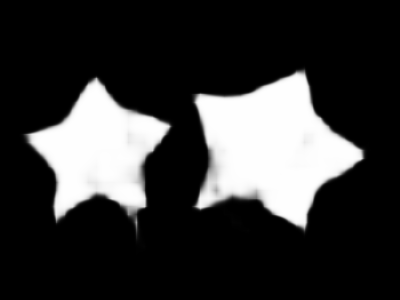}
     \end{minipage}
    \begin{minipage} [t] {0.12\linewidth}
        \centering
        \includegraphics [width=1\linewidth,height=0.75\linewidth] {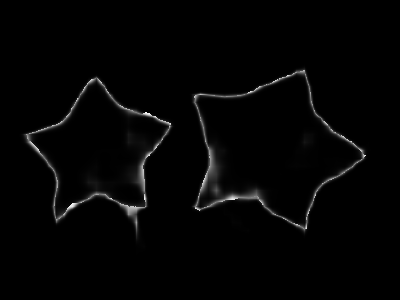}
    \end{minipage}
    \begin{minipage} [t] {0.12\linewidth}
        \centering
        \includegraphics [width=1\linewidth,height=0.75\linewidth] {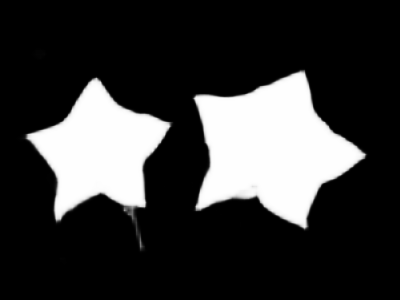}
     \end{minipage}    
    \begin{minipage} [t] {0.12\linewidth}
        \centering
        \includegraphics [width=1\linewidth,height=0.75\linewidth] {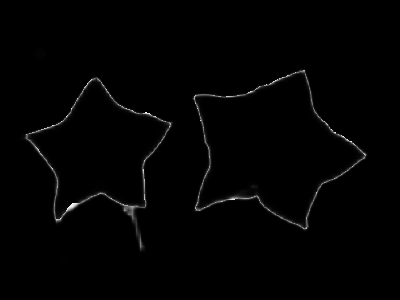}
    \end{minipage}         
    \begin{minipage} [t] {0.12\linewidth}
        \centering
        \includegraphics [width=1\linewidth,height=0.75\linewidth] {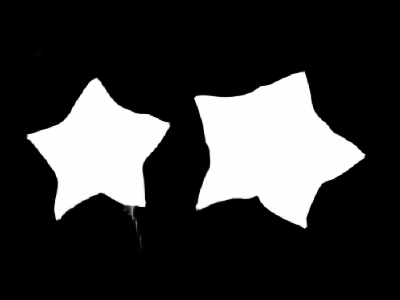}
     \end{minipage}    
    \begin{minipage} [t] {0.12\linewidth}
        \centering
        \includegraphics [width=1\linewidth,height=0.75\linewidth] {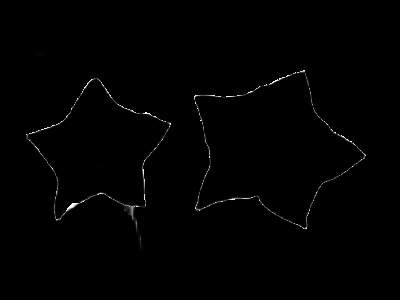}
    \end{minipage}     

    \begin{minipage} [t] {0.12\linewidth}
        \centering
        \includegraphics [width=1\linewidth,height=0.75\linewidth] 
        {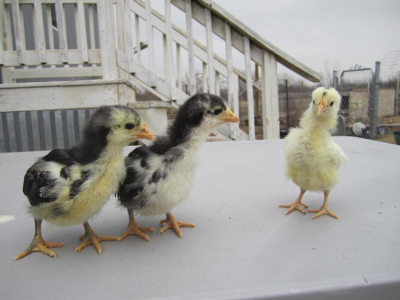}
    \end{minipage}
    \begin{minipage} [t] {0.12\linewidth}
        \centering
        \includegraphics [width=1\linewidth,height=0.75\linewidth] {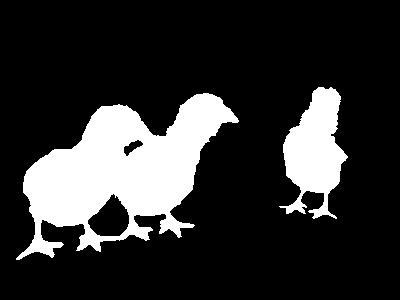}
    \end{minipage}
    \begin{minipage} [t] {0.12\linewidth}
        \centering
        \includegraphics [width=1\linewidth,height=0.75\linewidth] {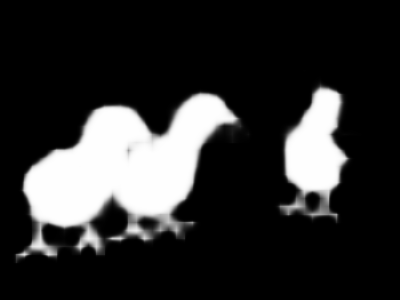}
     \end{minipage}
    \begin{minipage} [t] {0.12\linewidth}
        \centering
        \includegraphics [width=1\linewidth,height=0.75\linewidth] {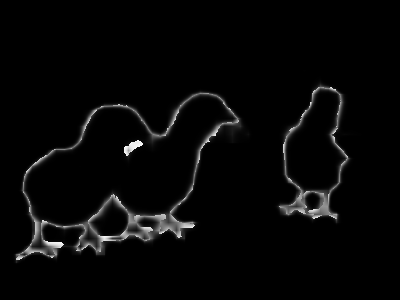}
    \end{minipage}
    \begin{minipage} [t] {0.12\linewidth}
        \centering
        \includegraphics [width=1\linewidth,height=0.75\linewidth] {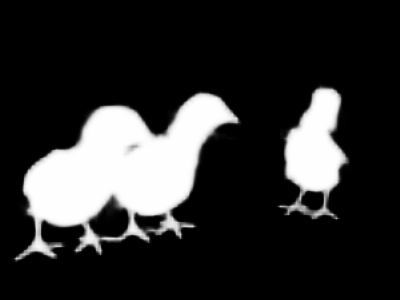}
     \end{minipage}    
    \begin{minipage} [t] {0.12\linewidth}
        \centering
        \includegraphics [width=1\linewidth,height=0.75\linewidth] {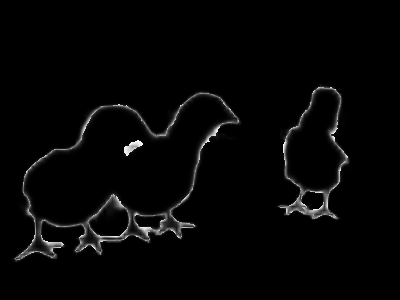}
    \end{minipage}        
    \begin{minipage} [t] {0.12\linewidth}
        \centering
        \includegraphics [width=1\linewidth,height=0.75\linewidth] {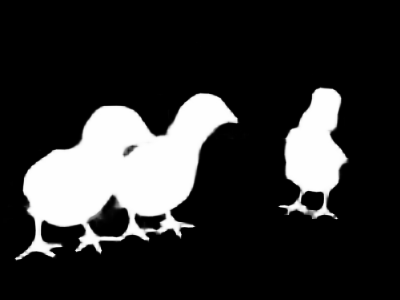}
     \end{minipage}    
    \begin{minipage} [t] {0.12\linewidth}
        \centering
        \includegraphics [width=1\linewidth,height=0.75\linewidth] {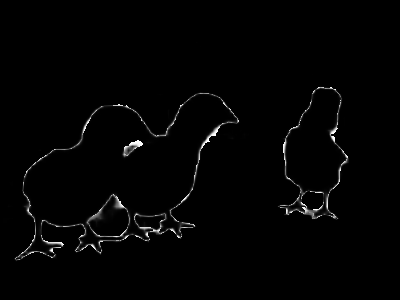}
    \end{minipage}    
    
    \begin{minipage} [t] {0.12\linewidth}
        \centering
        \includegraphics [width=1\linewidth,height=0.75\linewidth] 
        {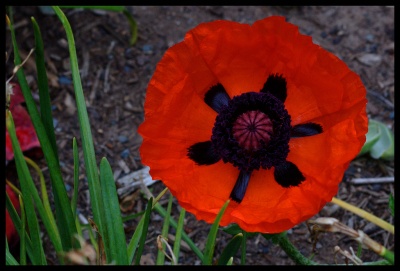}
        {(a)}
    \end{minipage}
    \begin{minipage} [t] {0.12\linewidth}
        \centering
        \includegraphics [width=1\linewidth,height=0.75\linewidth] {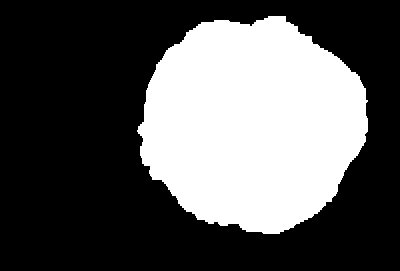}
        {(b)}
    \end{minipage}
    \begin{minipage} [t] {0.12\linewidth}
        \centering
        \includegraphics [width=1\linewidth,height=0.75\linewidth] {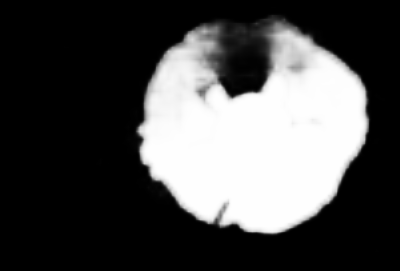}
        {(c)}
     \end{minipage}
    \begin{minipage} [t] {0.12\linewidth}
        \centering
        \includegraphics [width=1\linewidth,height=0.75\linewidth] {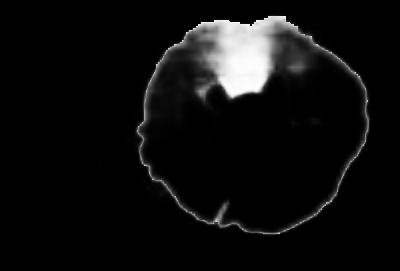}
        {(d)}
    \end{minipage}
    \begin{minipage} [t] {0.12\linewidth}
        \centering
        \includegraphics [width=1\linewidth,height=0.75\linewidth] {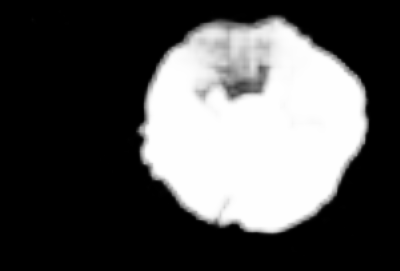}
        {(e)}
     \end{minipage}    
    \begin{minipage} [t] {0.12\linewidth}
        \centering
        \includegraphics [width=1\linewidth,height=0.75\linewidth] {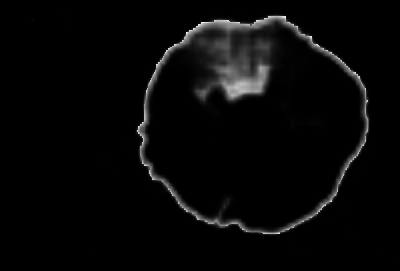}
        {(f)}
    \end{minipage}     
    \begin{minipage} [t] {0.12\linewidth}
        \centering
        \includegraphics [width=1\linewidth,height=0.75\linewidth] {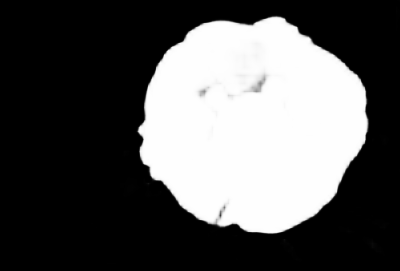}
        {(g)}
     \end{minipage}    
    \begin{minipage} [t] {0.12\linewidth}
        \centering
        \includegraphics [width=1\linewidth,height=0.75\linewidth] {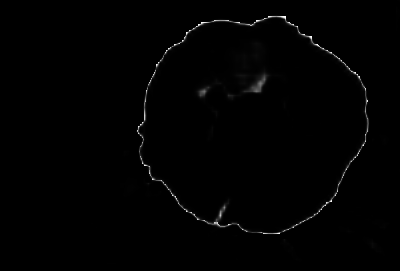}
        {(h)}
    \end{minipage}   
    \vspace{-0.2cm}
    \caption{Samples of the predictions and error maps. (a) is input image. (b) is ground truth. (c) is saliency map of \textbf{baseline}. (d) is error map of (c). (e) is saliency map of  \textbf{baseline+Grad} \cite{nonlocalsalient}. (f) is error map of (e). (g) is saliency map of  \textbf{baseline+BRL}. (h) is error map of (g).}
    \label{fig:error_map}
     \vspace{-0.1cm}
\end{figure}

As boundary causes errors on saliency prediction as Figure \ref{fig:error_map} shows, recent work such as \cite{contour} and \cite{boundary_aware} added a boundary (or contour) detection branch to the framework to enhance foreground mask generator. However, these structures bring additional parameters and computation and make the system more complex. In addition, because of the fragility of boundary prediction, recent SE2Net\cite{se2net} applied fully connected conditional random field (CRF) \cite{CRF} on both salient region prediction and boundary prediction to refine results before post-processing. Performing CRF twice is quite time-consuming. 

Differently, we propose a simple and yet effective refining strategy called Boundary Refinement Loss (BRL) on the existing segmentation branch. BRL only needs generating a loss on predictions, which involves negligible computation and no additional parameter. An efficient boundary refinement strategy is also proposed in \cite{nonlocalsalient} that utilizes gradients (denoted as `Grad') to generate the IoU Loss. We note that the boundary mask generated by gradients is not stable since the boundary is composed of a lot of values between 0 and 1 while this method only covers a small portion of the boundary area. The loss generating area of BRL covers both foreground and background, which is more robust than that of \cite{nonlocalsalient}. Therefore, such a strategy of \cite{nonlocalsalient} performs less well than BRL, as shown in Table \ref{tab:ablation_bound_layer}. 

As shown in Figure \ref{fig:error_map}(g)-(h), results refined by the proposed BRL have a much sharper boundary than baseline in (c)-(d) and `Grad'\cite{nonlocalsalient} in (e)-(f). Even the paws can be clearly detected in (g). Additionally, as shown in the last row of Figure \ref{fig:error_map}, the internal integrity of the salient region is better maintained if the boundary is refined well.

\begin{figure}
    \centering
    \begin{minipage} [t] {0.23\linewidth}
        \centering
        \includegraphics [width=1\linewidth,height=0.75\linewidth] {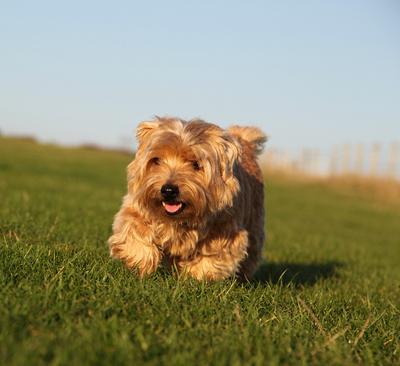}
        {(a) Input Image}
    \end{minipage}
    \begin{minipage} [t] {0.23\linewidth}
        \centering
        \includegraphics [width=1\linewidth,height=0.75\linewidth] {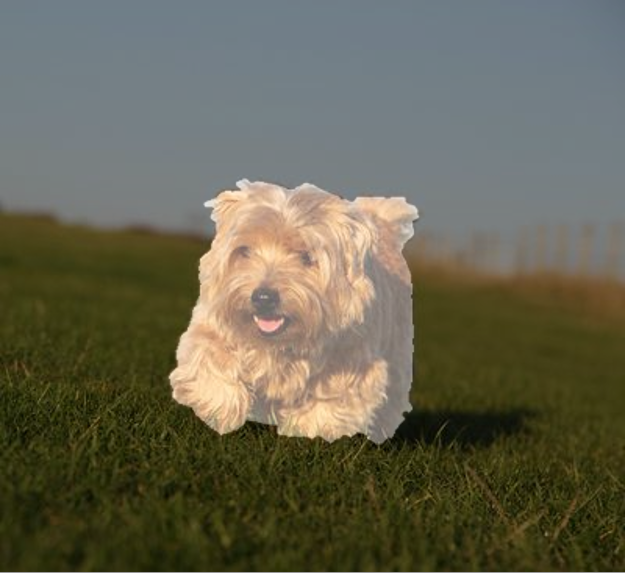}
        {(b) Ground Truth}
    \end{minipage}
    \begin{minipage} [t] {0.23\linewidth}
        \centering
        \includegraphics [width=1\linewidth,height=0.75\linewidth] {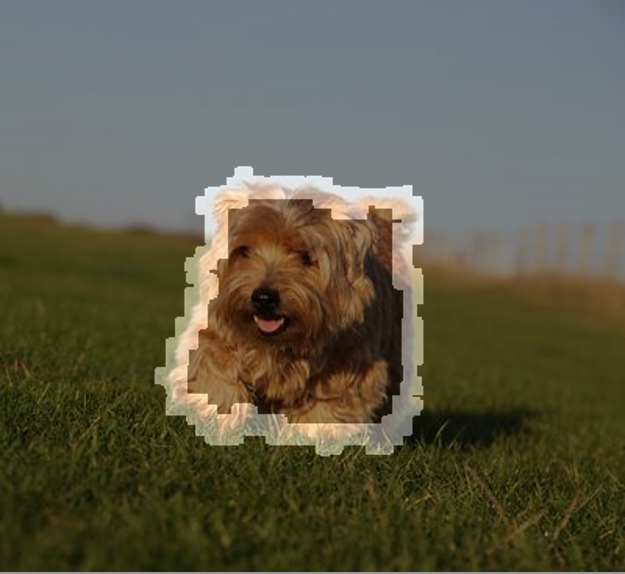}
        {(c) Boundary Area}
     \end{minipage}
     \vspace{-0.2cm}
    \caption{Example of boundary area generation.}
    \label{fig:boundary}
    \vspace{-0.5cm}
\end{figure}

\vspace{-0.2cm}
\subsubsection{Loss Formulation}
\vspace{-0.1cm}
 Before training, we generate a boundary area for each image as shown in Figure \ref{fig:boundary}. Since the width of the gradient boundary generated by Sobel operator is very small, we extract the boundary mask as shown in Figure \ref{fig:boundary}(c). Note that each pixel belonging to the original boundary expands on its surroundings to form the mask in (c). Therefore the new boundary area covers both background and foreground. We denote the expanding distance of each pixel as the \textit{Width} of the boundary area. We find \textit{Width}$=5$ yields best results and set it as the default without specification. \\
During training, BRL simply adds an extra boundary loss $L_{b}$ onto segmentation loss in the form of
\begin{equation}
\label{eqn:boundary_loss}
L_{b} = \mathcal{T}(P \odot B, G \odot B)
\end{equation}
where $\odot$ denotes pixel-wise multiplication, $P$ is the foreground prediction, $G$ is the ground truth, $B$ is the boundary area generated before training as Figure \ref{fig:boundary}(c) and $\mathcal{T}$ is the boundary loss function. In Eq. \eqref{eqn:boundary_loss}, $G$ and $P$ are masked by $B$, which filters out other areas and lets the loss function focus only on the foreground and background segmentation in the new boundary area. Similar to previous work \cite{nonlocalsalient}, we choose the IoU Loss as $\mathcal{T}$.
\vspace{-0.1in}
\begin{equation}
\label{eqn:final_loss}
L =L_f + \sigma\sum_{i=1}^{N}L_i + \eta (L_{bf} + \sigma\sum_{i=1}^{N}L_{bi})
\end{equation}
If BRL is used, the total loss $L$ is shown in Eq. \eqref{eqn:final_loss} where $L_{bf}$ and $L_{bi}$ are boundary losses of $\ma{Y}_{f}$ and $\ma{Y}_{i} (i = 1,2,......,N)$, and $\eta$ is used for weighing the sum of $L_{bf}$ and $L_{bi}$. Following \cite{nonlocalsalient}, we set $\eta$ to 1.0 in all experiments. As shown later in experiments, BRL improves the results by a large margin by refining the boundary area on segmentation prediction.

\vspace{-0.2cm}
\subsection{Region Refinement Network}
\begin{figure}
\centering
    \begin{minipage} [t] {0.6\linewidth}
        \centering
        \includegraphics [width=1.0\linewidth,height=0.6\linewidth] {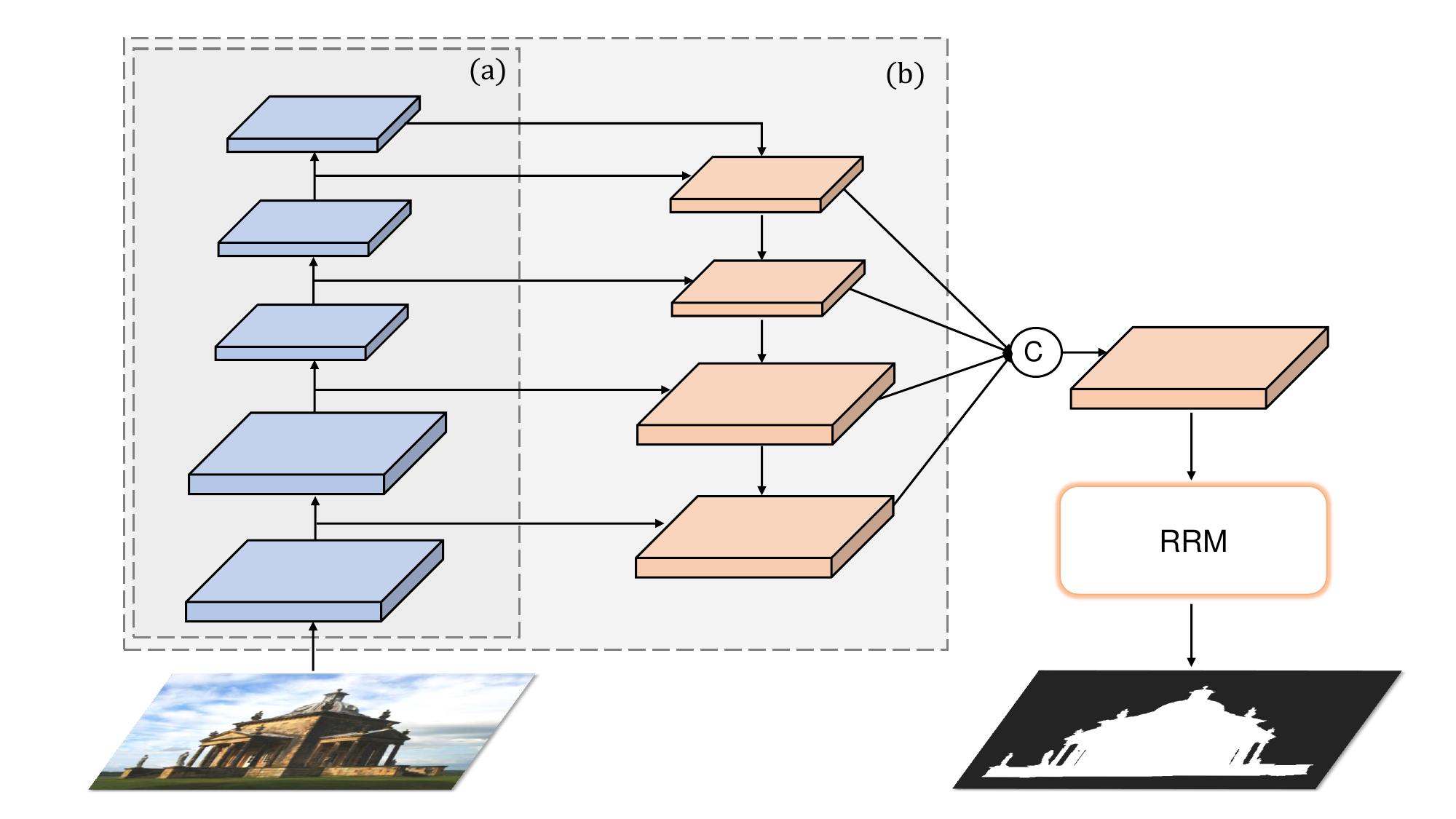}
    \end{minipage}
    \vspace{-0.2cm}
    \caption{Network Structure. (a) is ResNet-50\cite{resnet}, (b) is FPN\cite{FPN}. RRM is our proposed Region Refinement Module. ``C" means feature concatenation followed by a pyramid pooling module.}
    \label{fig:Network}
    \vspace{-0.5cm}
\end{figure}

\vspace{-0.1cm}
\label{sec:network}
To prove the effectiveness of RRM and BRL, we proposed the Region Refinement Network (RRN) as shown in Figure \ref{fig:Network}. The network contains a base network (b) and a Region Refinement Module (RRM). The base network sends features extracted from input images to RRM for further refinement, and BRL is then applied to the outputs of RRM.

Our baseline model (Figure \ref{fig:Network}(b)) is built upon ResNet-50\cite{resnet} (Figure \ref{fig:Network}(a)). In \cite{panet}, it is observed that {features in multiple levels together are helpful for accurate prediction}. We accordingly modify the network in the FPN\cite{FPN} style, as shown in Figure \ref{fig:Network}(b). By fusing features from four intermediate layers from FPN to form a multilevel feature before sending to the pyramid pooling module, the new base network outperforms the original PSPNet\cite{zhao2017pspnet} by a large margin, as demonstrated in Table \ref{tab:ablation_bound_layer}. The fusion of feature maps in different sizes is done by bilinear interpolation and concatenation.

For a fair comparison, we select VGG16 \cite{vgg}, ResNet-50 \cite{resnet} and ResNet-101 \cite{resnet} as our backbones.  
As for the VGG16 model, we extract features from Conv2\_2, Conv3\_3, Conv4\_3 and Conv5\_3 and send them to the FPN part to form a merged feature with $1/4$ of the resolution. For models based on ResNet-50 and 101, we use the same backbone configuration as PSPNet \cite{zhao2017pspnet}, and extract the last feature maps of the first MaxPool, Conv2, Conv3, Conv4 and Conv5 to the fusion stage. Features extracted from the backbone network are processed by 1$\times$1 convolutions with 128 output channels.

\vspace{-0.3cm}
\section{Experiments}
\vspace{-0.cm}

\vspace{-0.2cm}
\subsection{Settings}
\vspace{-0.1cm}
Our framework is based on PyTorch. The backbone network is initialized on VGG16, ResNet-50 and ResNet-101 trained on ImageNet. Other layers are initialized with PyTorch default setting. We adopt SGD with momentum as the optimizer where momentum and weight decay are set to 0.9 and 0.0001 respectively. Following \cite{deeplab}, we use the `poly' learning rate decay policy where current learning rate equals to the base learning rate multiplying $(1 - \frac{current_{iter}}{max_{iter}})^{power}$. We set the base rate to 0.01 and power to 0.9. As for hyper-parameters, we set $N$, $M$, $Width$, $\sigma$ and $\eta$ to 4, 3, 5, 2.0 and 1.0 respectively. Data augmentation is important for over-fitting prevention. In our experiments, we first perform mirroring and re-scaling from 0.75 and 1.25, and then add random rotation from -10 to 10 degrees on training images. Finally, we randomly crop $401\times401$ patches from the processed images as training samples. The batch size is 16, and we run 50 epochs for training in our setting. We output the saliency map from our network without additional refinement like the fully connected conditional random field (CRF) \cite{CRF} used in \cite{DCL}, \cite{DSS}, \cite{r3net} and \cite{liu2018PicaNet}.
Our experiments are conducted on an NVIDIA Tesla P40 GPU and Intel(R) Xeon(R) CPU E5-2680 v4 @ 2.40GHz.
\vspace{-0.1in}
\paragraph{Datasets}
We use the training data of DUTS \cite{DUTS} (10,553 images) for training as previous work \cite{cheng2019DNA},\cite{wang2018BiDirectional},\cite{wang2018BiDirectional},\cite{liu2018PicaNet}. Then we evaluate our model on HKU-IS \cite{HKU-IS} (4,447 images), 
ECSSD \cite{ECSSD} (1,000 images),
DUTS \cite{DUTS} (5,018 images), 
DUT-OMRON \cite{OMRON} (5,168 images) 
and PASCAL-S \cite{PASCAL-S} (850 images).
\zttian{For a fair comparison, we also train our model (VGG16) on MSRA-10K\cite{contrast1}(10000 images) and MSRA-B\cite{msra-b}(2500 images).}

\vspace{-0.2cm}
\paragraph{Evaluation Metrics}
Our evaluation metrics are Mean Absolute Error ($MAE$) and maximum F-measure ($F_{\beta}$). $MAE$ is the average pixel-wise error between prediction and ground truth as 
\begin{equation}
\label{eqn:MAE}
MAE = \frac{1}{H W}\sum_{i=1}^{H}\sum_{j=1}^{W}|P(i,j) - G(i,j)|
\end{equation}
where $P$ is prediction, $G$ is ground truth, $H$ and $W$ are height and width of the testing image. 
$F_{\beta}$ is used for measuring the overall performance \cite{ReverseAtten} with a balancing factor $\beta$ as
\begin{equation}
\label{eqn:MAE}
F_{\beta} = \frac{(1+\beta^2)(Recall + Precision)}{\beta^{2}Precision + Recall}
\end{equation}
where $\beta^2$ is set to 0.3 as suggested in \cite{SOD-Benchmark}. A good result usually yields a large $F_{\beta}$ and a small $MAE$.
\vspace{-0.2cm}
\subsection{Comparison with the State of the Art}
\vspace{-0.2cm}
\begin{table}[t]
    \renewcommand\arraystretch{5} 
    \scriptsize
    \centering
    \tabcolsep=0.2cm
    {
        \begin{tabular}{ c | c | c  c  c  c  c  c  c  c  c  c }
            \toprule
            \multirow{2}{*}{\textit{Method}} & 
            \multirow{2}{*}{\textit{Training}} & 
            \multicolumn{2}{c}{ECSSD} & 
            \multicolumn{2}{c}{HKU-IS} & 
            \multicolumn{2}{c}{OMRON} &
            \multicolumn{2}{c}{DUTS} &
            \multicolumn{2}{c}{PASCAL-S}\\
             & & \textit{F$_{\beta}$}$\uparrow$ & \textit{MAE}$\downarrow$ & \textit{F$_{\beta}$}$\uparrow$ & \textit{MAE}$\downarrow$ & \textit{F$_{\beta}$}$\uparrow$ & \textit{MAE}$\downarrow$ & \textit{F$_{\beta}$}$\uparrow$ & \textit{MAE}$\downarrow$ & \textit{F$_{\beta}$}$\uparrow$ & \textit{MAE}$\downarrow$ \\
            \specialrule{0em}{0pt}{1pt}
            \hline
            \specialrule{0em}{1pt}{0pt}
            \multicolumn{12}{c}{VGG-16 Backbone} \\
            \specialrule{0em}{0pt}{1pt}
            \hline
            \specialrule{0em}{1pt}{0pt}
            Amulet$_{2017}$ \cite{amulet} & MS-10K & 0.918 & 0.059 & 0.912 & 0.051 & 0.824 & 0.098 & 0.845 & 0.084 & 0.871 & 0.099  \\
            C2S-Net$_{2018}$ \cite{contour} & MS-10K 
            & 0.911 & 0.059 
            & 0.900 & 0.051 
            & 0.791 & {\color{red}\textbf{0.079}} 
            & 0.844 & 0.066
            & 0.865 & {\color{red}\textbf{0.086}}  \\ 
            \textbf{Ours} & MS-10K
            & {\color{red}\textbf{0.921}} & {\color{red}\textbf{0.052}}
            & {\color{red}\textbf{0.922}} & {\color{red}\textbf{0.041}} 
            & {\color{red}\textbf{0.827}} & 0.090 
            & {\color{red}\textbf{0.853}} & {\color{red}\textbf{0.064}}
            & {\color{red}\textbf{0.873}} & 0.103  \\ 
            \specialrule{0em}{0pt}{1pt}
            \hline
            \specialrule{0em}{1pt}{0pt}
            
            NLDF$_{2017}$ \cite{nonlocalsalient} & MS-B & 0.910 & 0.063 & 0.909 & 0.048 & 0.798 & 0.079 & 0.842 & 0.065 & 0.857 & 0.098  \\
            Chen \etal$_{2018}$ \cite{ReverseAtten} & MS-B & 0.906 & 0.056 & 0.898 & 0.045 & 0.807 & {\color{red}\textbf{0.062}} & 0.851 & 0.059 & 0.820 & 0.101  \\ 
            \textbf{Ours} & MS-B
            & {\color{red}\textbf{0.925}}& {\color{red}\textbf{0.051}}
            & {\color{red}\textbf{0.922}} & {\color{red}\textbf{0.039}}
            & {\color{red}\textbf{0.812}} & 0.073
            & {\color{red}\textbf{0.858}} & {\color{red}\textbf{0.057}} 
            & {\color{red}\textbf{0.859}} & {\color{red}\textbf{0.097}}  \\              
            \specialrule{0em}{0pt}{1pt}
            \hline
            \specialrule{0em}{1pt}{0pt}
            PAGR$_{2018}$ \cite{wang2018ProgAtten}$\dagger$ & DUTS & 0.911 & 0.061 & 0.904 & 0.047 & 0.768 & 0.071 & 0.849 & 0.055 & 0.851 & 0.089  \\  
            BMPM$_{2018}$ \cite{wang2018BiDirectional} & DUTS & 0.929 & 0.045 & 0.922 & 0.039 & 0.800 & 0.064 & 0.877 & 0.048 & 0.875 & 0.074  \\  
            PiCANet$_{2018}$ \cite{liu2018PicaNet} & DUTS & 0.935 & 0.047 & 0.924 & 0.042 & {\color{red}\textbf{0.833}} & 0.068 & 0.878 & 0.053 & {\color{red}\textbf{0.886}} & 0.078  \\   
            CPD$_{2019}$ \cite{cascadesodcvpr2019} & DUTS & 0.928 & 0.041 & 0.915 & 0.033 & 0.813 & {\color{red}\textbf{0.057}} & 0.877 & 0.043 & 0.876 & 0.073  \\  
            \textbf{Ours} & DUTS & {\color{red}\textbf{0.936}} & {\color{red}\textbf{0.040}} & {\color{red}\textbf{0.932}} & {\color{red}\textbf{0.031}} & 0.809 & 0.060 & {\color{red}\textbf{0.884}} & {\color{red}\textbf{0.040}} & {\color{red}\textbf{0.886}} & {\color{red}\textbf{0.072}}\\    
            
            \specialrule{0em}{0pt}{1pt}
            \hline
            \specialrule{0em}{1pt}{0pt}
            \multicolumn{12}{c}{ResNet-50 Backbone} \\
            \specialrule{0em}{0pt}{1pt}
            \hline
            \specialrule{0em}{1pt}{0pt}
            SRM$_{2017}$ \cite{srm} & DUTS & 0.923 & 0.054 & 0.913 & 0.046 & 0.812 & 0.069 & 0.857 & 0.058 & 0.870 & 0.084  \\
            DGRL$_{2018}$ \cite{exp6}& DUTS & 0.931 & 0.046 & 0.918 & 0.041 & 0.821 & 0.066 & 0.868 & 0.053 & 0.873 & 0.082  \\    
            PiCANet$_{2018}$ \cite{liu2018PicaNet}& DUTS & 0.938 & 0.047 & 0.923 & 0.043 & {\color{red}\textbf{0.840}}& 0.065 & 0.886 & 0.050 & 0.889 & 0.075  \\  
            
            CPD-R$_{2019}$ \cite{cascadesodcvpr2019}& DUTS & 0.937 & 0.037 & 0.916 & 0.034 & 0.826 & 0.056 & 0.878 & 0.043 & 0.881 & 0.071  \\     
            
            \textbf{Ours}& DUTS & {\color{red}\textbf{0.946}} & {\color{red}\textbf{0.034}} & {\color{red}\textbf{0.941}} & {\color{red}\textbf{0.027}} & 0.833 & {\color{red}\textbf{0.054}} & {\color{red}\textbf{0.893}} & {\color{red}\textbf{0.038}} & {\color{red}\textbf{0.900}} & {\color{red}\textbf{0.067}}\\ 
            
            \specialrule{0em}{0pt}{1pt}
            \hline
            \specialrule{0em}{1pt}{0pt}
            \multicolumn{12}{c}{ResNet-101 Backbone} \\
            \specialrule{0em}{0pt}{1pt}
            \hline
            \specialrule{0em}{1pt}{0pt}
            R3Net-C$_{2018}$\cite{r3net}$\ddagger$ & MS-10K & 0.934 & 0.040 & 0.922 & 0.036 & 0.841 & 0.063 & 0.866 & 0.057 & 0.841 & 0.092  \\  
            R3Net-C$_{2018}$\cite{r3net}$\ddagger*$ & DUTS
            & 0.943 & 0.042
            & 0.932 & 0.038 
            & 0.840 & 0.065 
            & 0.896 & 0.045 
            & 0.895 & 0.072  \\

            \textbf{Ours}& DUTS & {\color{red}\textbf{0.954}} & {\color{red}\textbf{0.030}} & {\color{red}\textbf{0.944}} & {\color{red}\textbf{0.025}} & {\color{red}\textbf{0.843}} & {\color{red}\textbf{0.052}} & {\color{red}\textbf{0.901}} & {\color{red}\textbf{0.036}} & {\color{red}\textbf{0.900}} & {\color{red}\textbf{0.063}} \\ 
            
            \bottomrule
        \end{tabular}
    }
    \vspace{0.1cm}
    \caption{Results on five benchmark datasets. `MS-10K': MSRA10K\cite{contrast1}, `MS-B': MSRA-B\cite{msra-b}. `DUTS': the training data of DUTS\cite{DUTS}. `C': results refined by CRF\cite{CRF}. $\dagger$ : VGG-19 backbone. $\ddagger$: ResNeXt-101\cite{resnext} backbone. $*$: Reproduced result. The best results are shown in {\color{red}\textbf{Red}}.}

    \vspace{-0.2cm}
    \label{tab:compare_sota}
\end{table}

\begin{figure}
    \begin{minipage} [t] {0.095\linewidth}
        \centering
        \includegraphics [width=1\linewidth,height=0.75\linewidth] {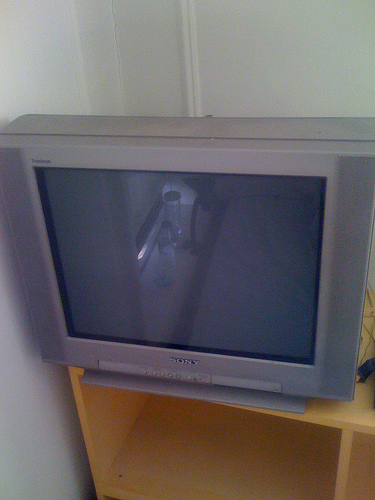}
    \end{minipage}
    \begin{minipage} [t] {0.095\linewidth}
        \centering
        \includegraphics [width=1\linewidth,height=0.75\linewidth] {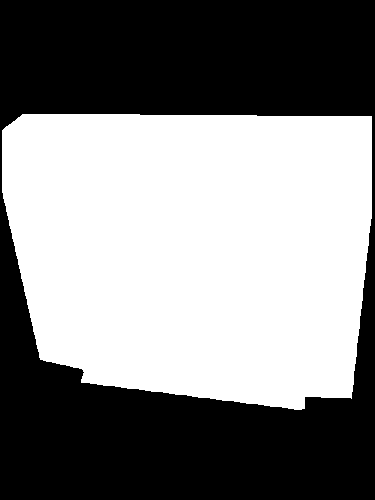}
    \end{minipage}
    \begin{minipage} [t] {0.095\linewidth}
        \centering
        \includegraphics [width=1\linewidth,height=0.75\linewidth] {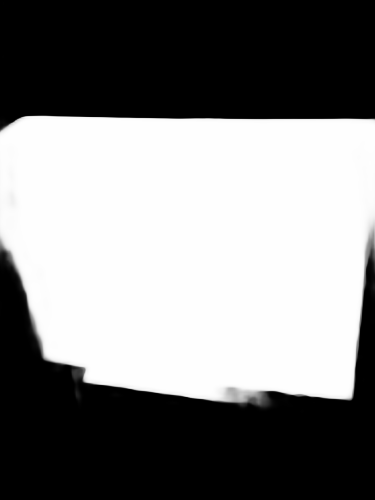}
    \end{minipage}
    \begin{minipage} [t] {0.095\linewidth}
        \centering
        \includegraphics [width=1\linewidth,height=0.75\linewidth] {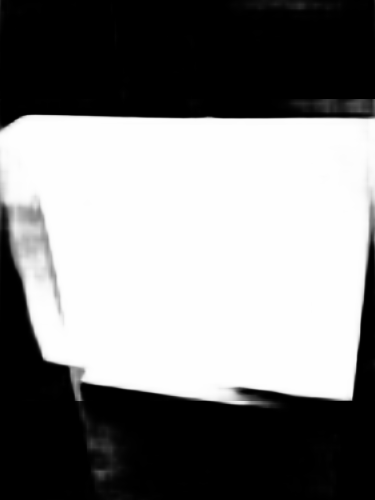}
    \end{minipage}    
    \begin{minipage} [t] {0.095\linewidth}
        \centering
        \includegraphics [width=1\linewidth,height=0.75\linewidth] {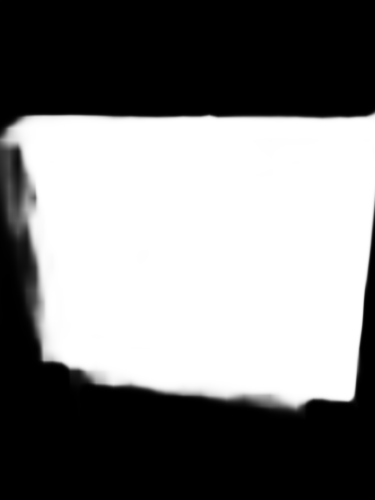}
    \end{minipage}   
    \begin{minipage} [t] {0.095\linewidth}
        \centering
        \includegraphics [width=1\linewidth,height=0.75\linewidth] {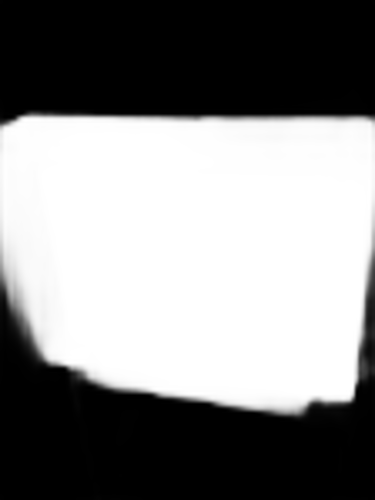}
    \end{minipage} 
    \begin{minipage} [t] {0.095\linewidth}
        \centering
        \includegraphics [width=1\linewidth,height=0.75\linewidth] {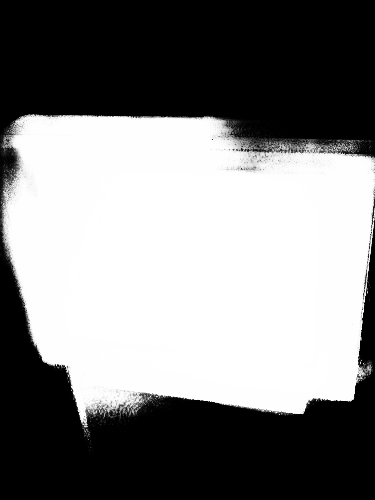}
    \end{minipage}      
    \begin{minipage} [t] {0.095\linewidth}
        \centering
        \includegraphics [width=1\linewidth,height=0.75\linewidth] {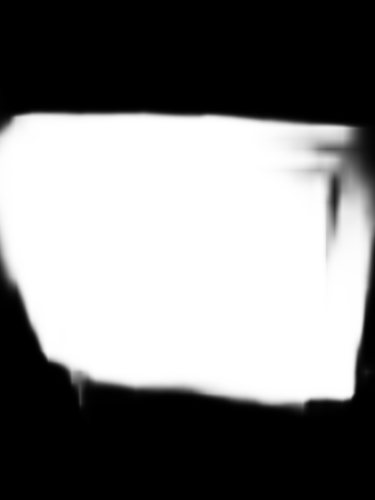}
    \end{minipage}          
    \begin{minipage} [t] {0.095\linewidth}
        \centering
        \includegraphics [width=1\linewidth,height=0.75\linewidth] {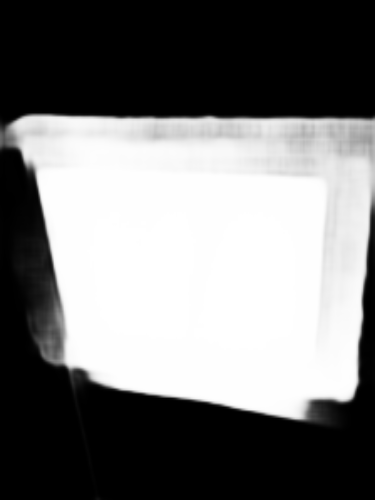}
    \end{minipage}    
    \begin{minipage} [t] {0.095\linewidth}
        \centering
        \includegraphics [width=1\linewidth,height=0.75\linewidth] {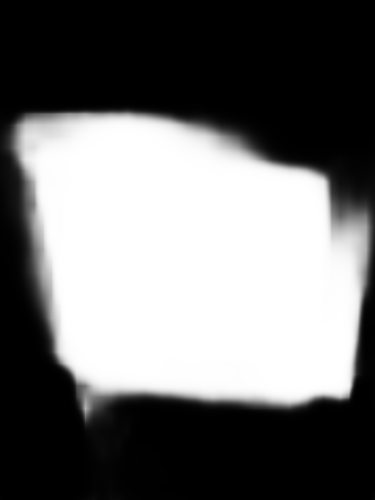}
    \end{minipage}  
    
    \begin{minipage} [t] {0.095\linewidth}
        \centering
        \includegraphics [width=1\linewidth,height=0.75\linewidth] {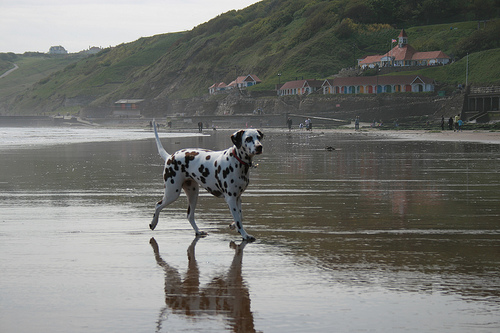}
    \end{minipage}
    \begin{minipage} [t] {0.095\linewidth}
        \centering
        \includegraphics [width=1\linewidth,height=0.75\linewidth] {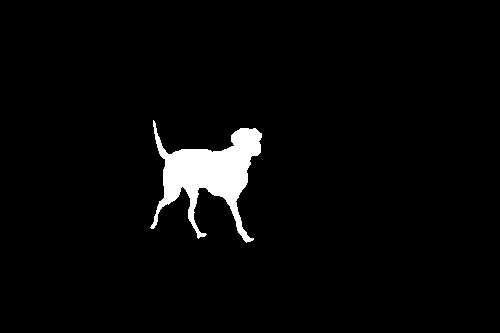}
    \end{minipage}
    \begin{minipage} [t] {0.095\linewidth}
        \centering
        \includegraphics [width=1\linewidth,height=0.75\linewidth] {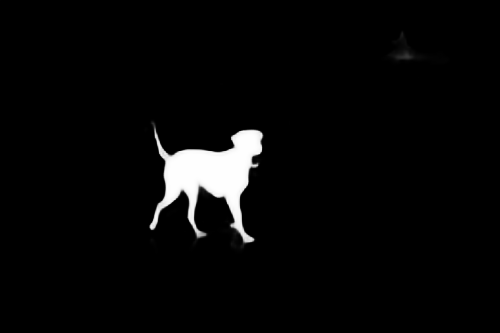}
    \end{minipage}
    \begin{minipage} [t] {0.095\linewidth}
        \centering
        \includegraphics [width=1\linewidth,height=0.75\linewidth] {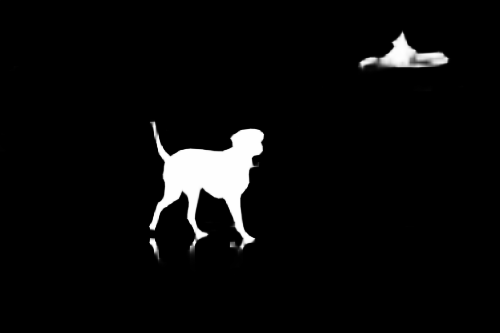}
    \end{minipage}     
    \begin{minipage} [t] {0.095\linewidth}
        \centering
        \includegraphics [width=1\linewidth,height=0.75\linewidth] {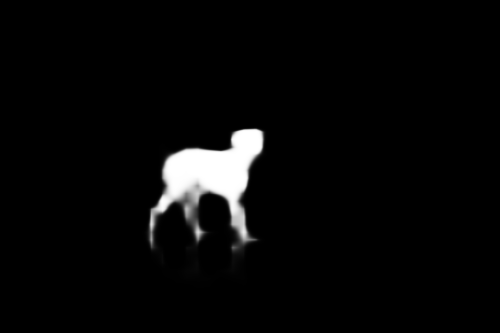}
    \end{minipage}   
    \begin{minipage} [t] {0.095\linewidth}
        \centering
        \includegraphics [width=1\linewidth,height=0.75\linewidth] {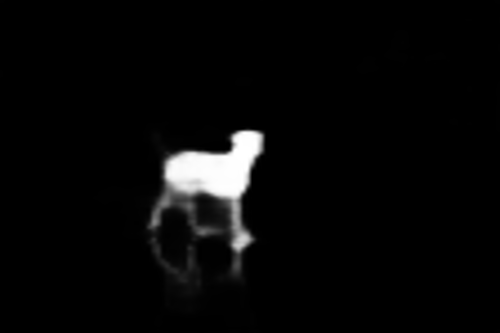}
    \end{minipage} 
    \begin{minipage} [t] {0.095\linewidth}
        \centering
        \includegraphics [width=1\linewidth,height=0.75\linewidth] {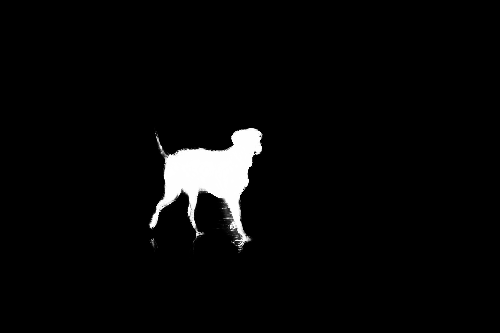}
    \end{minipage}      
    \begin{minipage} [t] {0.095\linewidth}
        \centering
        \includegraphics [width=1\linewidth,height=0.75\linewidth] {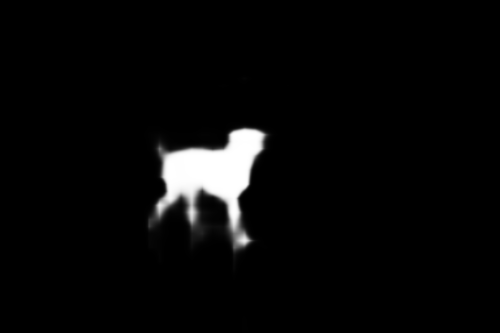}
    \end{minipage}          
    \begin{minipage} [t] {0.095\linewidth}
        \centering
        \includegraphics [width=1\linewidth,height=0.75\linewidth] {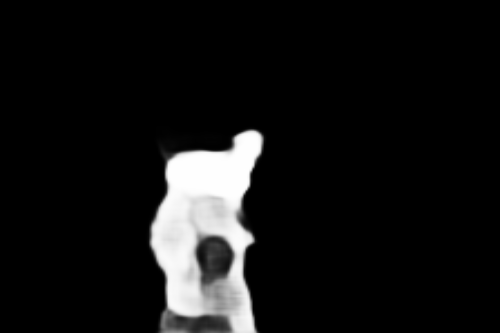}
    \end{minipage}    
    \begin{minipage} [t] {0.095\linewidth}
        \centering
        \includegraphics [width=1\linewidth,height=0.75\linewidth] {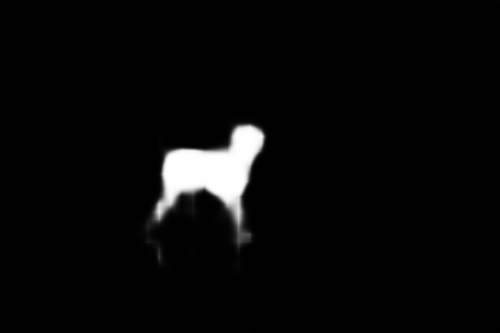}
    \end{minipage}  
    
    \begin{minipage} [t] {0.095\linewidth}
        \centering
        \includegraphics [width=1\linewidth,height=0.75\linewidth] {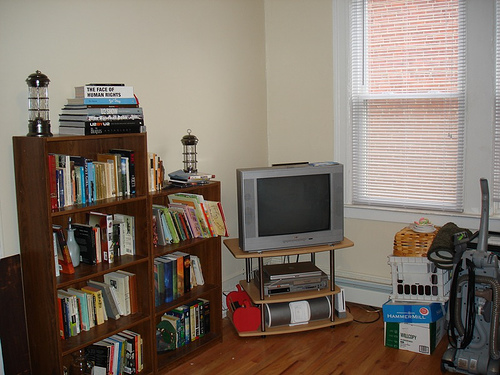}
    \end{minipage}
    \begin{minipage} [t] {0.095\linewidth}
        \centering
        \includegraphics [width=1\linewidth,height=0.75\linewidth] {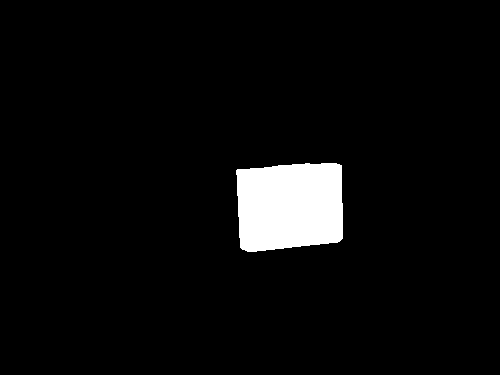}
    \end{minipage}
    \begin{minipage} [t] {0.095\linewidth}
        \centering
        \includegraphics [width=1\linewidth,height=0.75\linewidth] {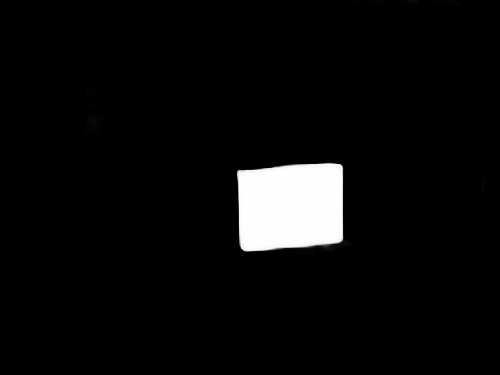}
    \end{minipage}
    \begin{minipage} [t] {0.095\linewidth}
        \centering
        \includegraphics [width=1\linewidth,height=0.75\linewidth] {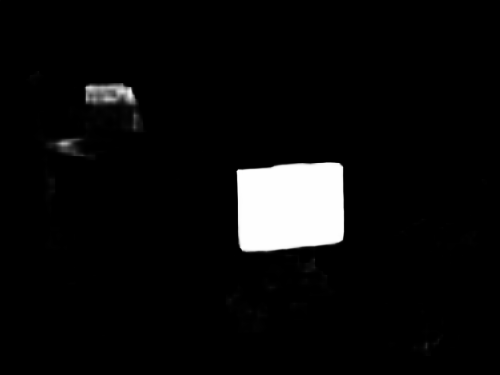}
    \end{minipage}    
    \begin{minipage} [t] {0.095\linewidth}
        \centering
        \includegraphics [width=1\linewidth,height=0.75\linewidth] {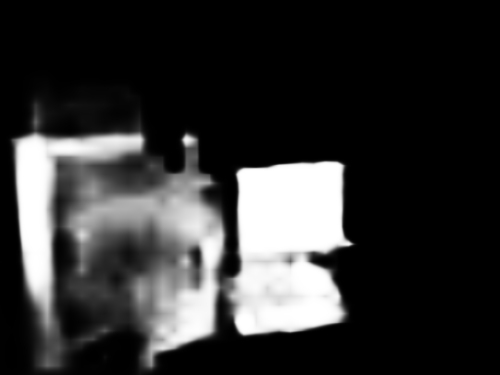}
    \end{minipage}   
    \begin{minipage} [t] {0.095\linewidth}
        \centering
        \includegraphics [width=1\linewidth,height=0.75\linewidth] {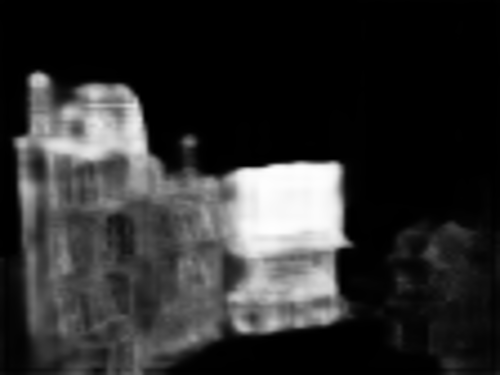}
    \end{minipage} 
    \begin{minipage} [t] {0.095\linewidth}
        \centering
        \includegraphics [width=1\linewidth,height=0.75\linewidth] {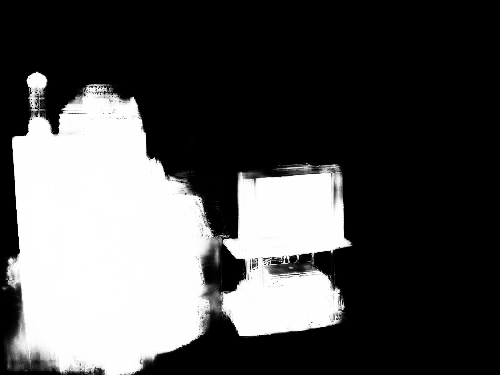}
    \end{minipage}      
    \begin{minipage} [t] {0.095\linewidth}
        \centering
        \includegraphics [width=1\linewidth,height=0.75\linewidth] {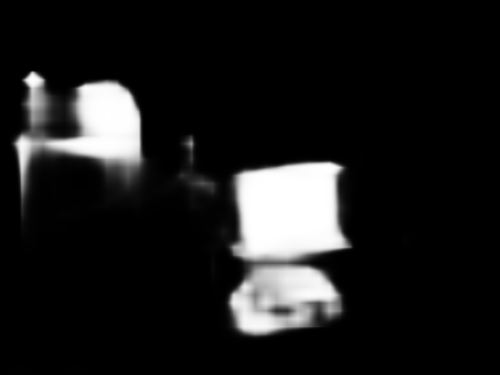}
    \end{minipage}          
    \begin{minipage} [t] {0.095\linewidth}
        \centering
        \includegraphics [width=1\linewidth,height=0.75\linewidth] {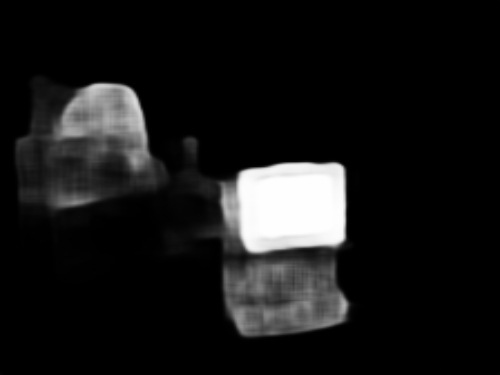}
    \end{minipage}    
    \begin{minipage} [t] {0.095\linewidth}
        \centering
        \includegraphics [width=1\linewidth,height=0.75\linewidth] {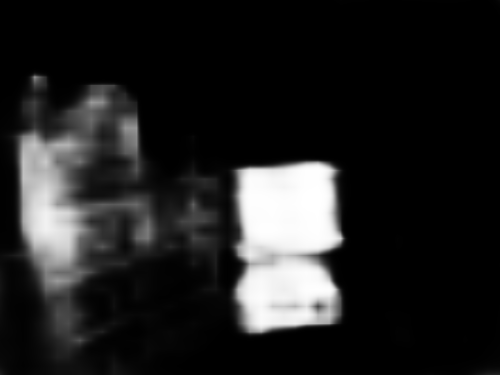}
    \end{minipage}  
    
    \begin{minipage} [t] {0.095\linewidth}
        \centering
        \includegraphics [width=1\linewidth,height=0.75\linewidth] {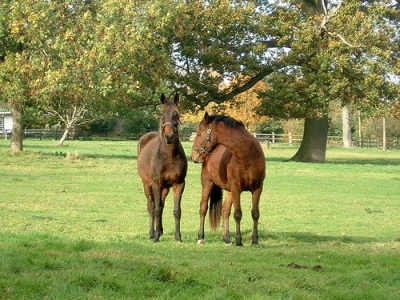}
        {(a)}
    \end{minipage}
    \begin{minipage} [t] {0.095\linewidth}
        \centering
        \includegraphics [width=1\linewidth,height=0.75\linewidth] {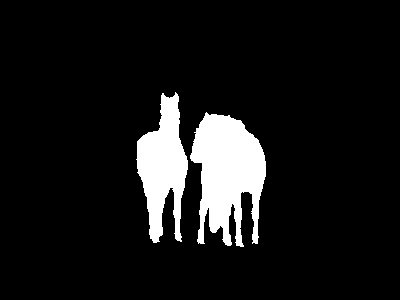}
        {(b)}
    \end{minipage}
    \begin{minipage} [t] {0.095\linewidth}
        \centering
        \includegraphics [width=1\linewidth,height=0.75\linewidth] {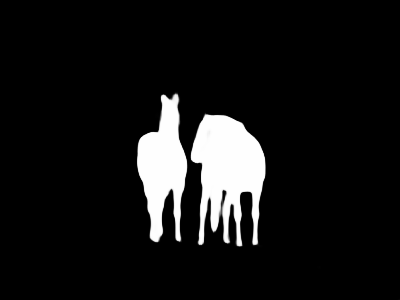}
        {(c)}
    \end{minipage}
    \begin{minipage} [t] {0.095\linewidth}
        \centering
        \includegraphics [width=1\linewidth,height=0.75\linewidth] {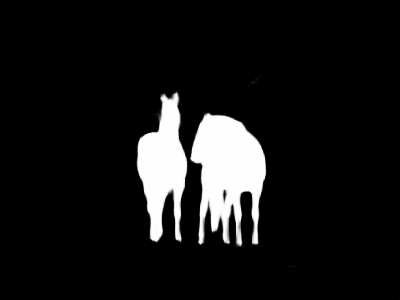}
        {(d)}
    \end{minipage}      
    \begin{minipage} [t] {0.095\linewidth}
        \centering
        \includegraphics [width=1\linewidth,height=0.75\linewidth] {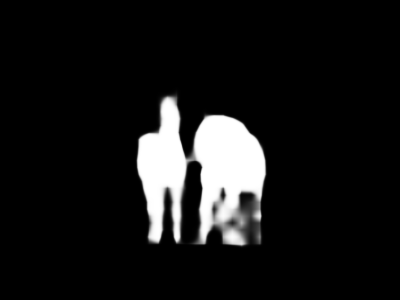}
        {(e)}
    \end{minipage}   
    \begin{minipage} [t] {0.095\linewidth}
        \centering
        \includegraphics [width=1\linewidth,height=0.75\linewidth] {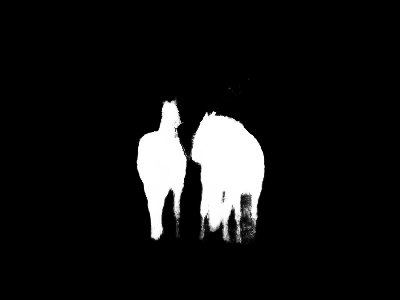}
        {(f)}
    \end{minipage} 
    \begin{minipage} [t] {0.095\linewidth}
        \centering
        \includegraphics [width=1\linewidth,height=0.75\linewidth] {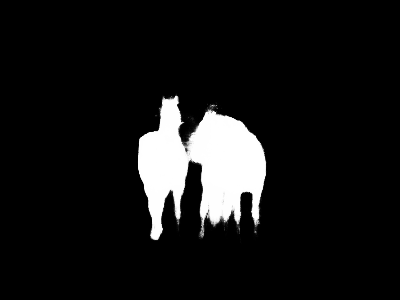}
        {(g)}
    \end{minipage}      
    \begin{minipage} [t] {0.095\linewidth}
        \centering
        \includegraphics [width=1\linewidth,height=0.75\linewidth] {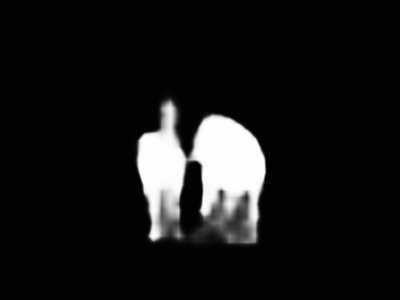}
        {(h)}
    \end{minipage}          
    \begin{minipage} [t] {0.095\linewidth}
        \centering
        \includegraphics [width=1\linewidth,height=0.75\linewidth] {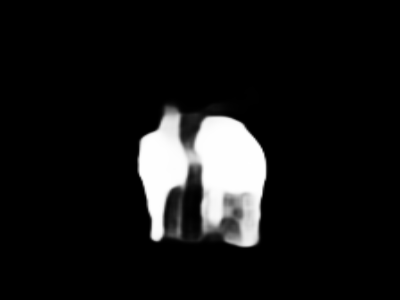}
        {(i)}
    \end{minipage}    
    \begin{minipage} [t] {0.095\linewidth}
        \centering
        \includegraphics [width=1\linewidth,height=0.75\linewidth] {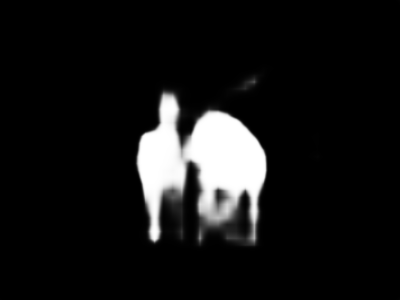}
        {(j)}
    \end{minipage} 
    \vspace{-0.3cm}
    \caption{Qualitative comparison with previous state-of-the-art methods. (a)  {Input image}. (b)  {Ground truth}. (c)  {Baseline+BRL+RRM}. (d) { Baseline+BRL}. 
    (e)  {CPD-R} \cite{cascadesodcvpr2019}. (f)  {PiCANet-R} \cite{liu2018PicaNet}. (g)  {R3-Net} \cite{r3net}. (h) {DGRL} \cite{exp6}. (i)  {C2S-Net} \cite{contour}.
    (j)  {SRM} \cite{srm}. Results of R3-Net are refined by CRF \cite{CRF}.}
    \label{fig:sample}
    \vspace{-0.3cm}
\end{figure}

In this section, we compare our proposed method with several state-of-the-art deep learning based algorithms shown in Table \ref{tab:compare_sota}. For a fair comparison, we calculate the results on the salient maps; or the models provided by authors with the same evaluation code. It shows that our method achieves the new state-of-the-art on all backbone methods. The speeds of our final models with VGG16, ResNet50 and ResNet101 are 54.8, 52.1 and 32.8 FPS respectively on $352\times352$ input.

For visual comparison, we compare the salient maps with state-of-the-art methods (CPD-R \cite{cascadesodcvpr2019},  PiCANet-R \cite{liu2018PicaNet}, R3-Net \cite{r3net}, DGRL \cite{exp6}, C2S-Net \cite{contour}, SRM \cite{srm}) shown in Figure \ref{fig:sample}. In Figure \ref{fig:sample}, our final model (c) is composed of both BRL and RRM. It is more robust to large objects, tiny structures like limbs and ears, complex background, and multiple objects, which prove its effectiveness. Additionally, our method yields much better boundaries in salient maps than other methods.

\vspace{-0.2cm}
\subsection{Ablation Study and Analysis}
\vspace{-0.1cm}
\subsubsection{Effectiveness of BRL}
\vspace{-0.1cm}
\label{sec:effect_BRL}
In order to demonstrate the effectiveness of our proposed Boundary Refinement Loss (BRL), we compare BRL with the enhancement strategy (denoted as `Grad') proposed in \cite{nonlocalsalient} and the baseline structure in Table \ref{tab:ablation_bound_layer}. `Grad' yields slight improvements over the baseline model except for PASCAL-S in terms of $F_{\beta}$ and $MAE$. Our proposed BRL makes the baseline model generate higher quality results on all benchmark datasets. The results are even comparable with previous state-of-the-art ones. The qualitative comparison is shown in Figure \ref{fig:error_map} where BRL in (g)-(h) refines boundary better than Grad \cite{nonlocalsalient} in (e)-(f). We put experiments on different boundary \textit{Width} in the supplementary file.

\begin{table}[t]
    \renewcommand\arraystretch{5} 
    \scriptsize
    \centering
    \tabcolsep=0.2cm
    {
        \begin{tabular}{ c | c  c  c  c  c  c  c  c  c  c }
            \toprule
            \multirow{2}{*}{\textit{Method}} & 
            \multicolumn{2}{c}{ECSSD} & 
            \multicolumn{2}{c}{HKU-IS} & 
            \multicolumn{2}{c}{OMRON} &
            \multicolumn{2}{c}{DUTS} &
            \multicolumn{2}{c}{PASCAL-S}\\
             & \textit{F$_{\beta}$}$\uparrow$ & \textit{MAE}$\downarrow$ & \textit{F$_{\beta}$}$\uparrow$ & \textit{MAE}$\downarrow$ & \textit{F$_{\beta}$}$\uparrow$ & \textit{MAE}$\downarrow$ & \textit{F$_{\beta}$}$\uparrow$ & \textit{MAE}$\downarrow$ & \textit{F$_{\beta}$}$\uparrow$ & \textit{MAE}$\downarrow$ \\
            \specialrule{0em}{0pt}{1pt}
            \hline
            \specialrule{0em}{1pt}{0pt}
            PSPNet\cite{zhao2017pspnet}
            & 0.930 & 0.048 & 0.928 & 0.039 & 0.820 & 0.066 & 0.886 & 0.046 & 0.877 & 0.081  \\            
            Baseline
            & 0.936 & 0.043 & 0.931 & 0.035 & 0.819 & 0.063 & 0.886 & 0.044 & 0.878 & 0.079  \\
            Baseline+Grad\cite{nonlocalsalient}
            & 0.939 & 0.042 & 0.935 & 0.033 & 0.824 & 0.062 & 0.888 & 0.044 & 0.877 & 0.080  \\
            Baseline+BRL
            & 0.943 & 0.038 
            & 0.934 & 0.030 
            & {\color{red}\textbf{0.833}} & 0.061 
            & 0.889 & 0.042 
            & 0.882 & 0.076  \\

            \textbf{Baseline+BRL+RRM}
            & {\color{red}\textbf{0.946}} & {\color{red}\textbf{0.034}} 
            & {\color{red}\textbf{0.941}} & {\color{red}\textbf{0.027}} 
            & {\color{red}\textbf{0.833}} & {\color{red}\textbf{0.054}} 
            & {\color{red}\textbf{0.893}} & {\color{red}\textbf{0.038}} 
            & {\color{red}\textbf{0.900}} & {\color{red}\textbf{0.067}}  \\
            \bottomrule
        \end{tabular}
    }
    \vspace{0.1cm}
    \caption{Comparisons of different boundary refinement strategies. The baseline is shown in Figure \ref{fig:Network}(b). All models use ResNet50 as the backbone.}
    \vspace{-0.3cm}
    \label{tab:ablation_bound_layer}
\end{table}

\begin{table}[t]
    \renewcommand\arraystretch{5} 
    \scriptsize
    \centering
    \tabcolsep=0.2cm
    {
        \begin{tabular}{ c |  c  c  c  c  c  c  c  c  c  c }
            \toprule
            \multirow{2}{*}{\textit{$\sigma$}} & 
            \multicolumn{2}{c}{ECSSD} & 
            \multicolumn{2}{c}{HKU-IS} & 
            \multicolumn{2}{c}{OMRON} &
            \multicolumn{2}{c}{DUTS} &
            \multicolumn{2}{c}{PASCAL-S}\\
             & \textit{F$_{\beta}$}$\uparrow$ & \textit{MAE}$\downarrow$ & \textit{F$_{\beta}$}$\uparrow$ & \textit{MAE}$\downarrow$ & \textit{F$_{\beta}$}$\uparrow$ & \textit{MAE}$\downarrow$ & \textit{F$_{\beta}$}$\uparrow$ & \textit{MAE}$\downarrow$ & \textit{F$_{\beta}$}$\uparrow$ & \textit{MAE}$\downarrow$ \\
            \specialrule{0em}{0pt}{1pt}
            \hline
            \specialrule{0em}{1pt}{0pt}
            0.00
            & 0.940 & 0.037 & 0.936 & 0.030 & 0.830 & 0.059 & 0.892 & 0.042 & 0.878 & 0.074  \\
            0.25
            & {\color{red}\textbf{0.947}} & {\color{red}\textbf{0.034}} & 0.940 & {\color{red}\textbf{0.027}} & 0.832 & 0.057 & {\color{red}\textbf{0.898}} & 0.039 & 0.891 & {\color{red}\textbf{0.070}}  \\
            \textbf{0.50}
            & 0.946 & {\color{red}\textbf{0.034}} & {\color{red}\textbf{0.941}} & {\color{red}\textbf{0.027}} & 0.833 & {\color{red}\textbf{0.054}} & 0.893 & {\color{red}\textbf{0.038}} & {\color{red}\textbf{0.900}} & {\color{red}\textbf{0.067}}   \\
            0.75
            & 0.946 & 0.035 & 0.939 & 0.028 & {\color{red}\textbf{0.839}} & 0.057 & 0.895 & 0.040 & 0.889 & 0.072  \\
            1.00
            & 0.944 & 0.036 & 0.940 & 0.028 & {\color{red}\textbf{0.839}} & 0.057 & 0.891 & 0.040 & 0.880 & 0.073  \\
            
            \bottomrule
        \end{tabular}
    }
    \caption{Comparison between results of different $\sigma$. The best results are shown in {{red}}. }
    \vspace{-0.6cm}
    \label{tab:sigma}
\end{table}

\begin{table}[H]
    \renewcommand\arraystretch{5} 
    \scriptsize
    \centering
    \tabcolsep=0.2cm
    {
        \begin{tabular}{ c |  c  c c  c  c  c  c  c  c  c }
            \toprule
            \multirow{2}{*}{\textit{Method}} & 
            \multicolumn{2}{c}{ECSSD} & 
            \multicolumn{2}{c}{HKU-IS} & 
            \multicolumn{2}{c}{OMRON} &
            \multicolumn{2}{c}{DUTS} &
            \multicolumn{2}{c}{PASCAL-S}\\
             & \textit{F$_{\beta}$}$\uparrow$ & \textit{MAE}$\downarrow$ & \textit{F$_{\beta}$}$\uparrow$ & \textit{MAE}$\downarrow$ & \textit{F$_{\beta}$}$\uparrow$ & \textit{MAE}$\downarrow$ & \textit{F$_{\beta}$}$\uparrow$ & \textit{MAE}$\downarrow$ & \textit{F$_{\beta}$}$\uparrow$ & \textit{MAE}$\downarrow$ \\
            \specialrule{0em}{0pt}{1pt}
            \hline
            \specialrule{0em}{1pt}{0pt}
            Mask
            & {\color{red}\textbf{0.946}} & {\color{red}\textbf{0.034}} & {\color{red}\textbf{0.941}} & {\color{red}\textbf{0.027}} & 0.833 & {\color{red}\textbf{0.054}} & 0.893 & {\color{red}\textbf{0.038}} & {\color{red}\textbf{0.900}} & {\color{red}\textbf{0.067}}  \\
            Non-Mask
            & 0.943 & 0.036 & 0.937 & 0.030 & {\color{red}\textbf{0.834}} & 0.058 & {\color{red}\textbf{0.894}} & 0.042 & 0.890 & 0.074  \\
            
            \bottomrule
        \end{tabular}
    }
    \caption{Comparisons of model with/without mask. $N=4, M=3$ and $\sigma = 0.5$ in two experiments.}
     \vspace{-0.4cm}
    \label{tab:mask}
\end{table}

\vspace{-0.4cm}
\subsubsection{Effectiveness of RRM}
\vspace{-0.1cm}
\label{sec:effect_grm}
As shown in Figure \ref{fig:sample}(c), RRM further refines the results by  reducing false prediction existing in Figure \ref{fig:sample}(d), and the model with RRM ($N=4,M=3$) achieves best performance in Table \ref{tab:ablation_bound_layer}. In Eq. \eqref{eqn:final_loss}, we follow \cite{nonlocalsalient} to set $\eta$ to 1.0. There are three parameters ($N, M$ and $\sigma$) that are not decided. 

In this section, we first conduct experiments on different values of $\sigma$ to find the best balancing factor. Then, we show the effectiveness of the mask used as attention for controlling information flow. Examples illustrating the refinement process in each stage of RRM and experiments on different values of $N$ (layer number) and $M$ (convolution number) are shown in the supplementary material. We find RRM with ($N=4, M=3$) brings the best performance.

\vspace{-0.35cm}
\paragraph{Balancing Factor ($\sigma$)}
The side outputs from intermediate layers of RRM generate auxiliary losses $\sigma\sum_{i=1}^{N}L_i$ that may influence the learning of final output. In Table \ref{tab:sigma}, to find the best $\sigma$ for balancing learning between side and final output, we select five values for $\sigma$ as [0, 0.25, 0.5, 0.75, 1.0] and find that 0.5 yields the best performance. When $\sigma = 0$, there is no supervision on side outputs, which causes self-learned attention maps. However, as shown in our experiments, the supervision of ground truth on side outputs produces better attention maps to control the information flow in RRM. Also, when $\sigma$ is close to one, the loss information back-propagated from the final output is dominated by previous layers' outputs; a large $\sigma$ yields sub-optimal final results. Therefore $\sigma$ should be close to 0.5 for optimal results shown in Table \ref{tab:sigma}.

\vspace{-0.35cm}
\paragraph{Mask in RRM}
In Figure \ref{fig:GRM}, we show the output of the previous layer is used as an attention mask applied to the next layer to control the information flow. We compare models with and without masks in RRM to evaluate the usefulness of the mask and show results in Table \ref{tab:mask}. It is clear that when $N$ and $\sigma$ are set reasonably, without the mask to control the useful information flow in the module, redundant information becomes the bottleneck.

\vspace{-0.2cm}
\subsection{Applications}
\vspace{-0.1cm}
Our method (RRN) can be applied to other tasks to further prove its effectiveness. For Portrait segmentation based on the portrait dataset \cite{portrait}, we apply RRN and make a comparison with the results of \cite{cascadesodcvpr2019}. As listed in Table \ref{tab:portrait}, RRN (VGG-16 backbone) achieves the best mIoU performance.

We also apply RRN (VGG-16 backbone) to shadow detection and evaluate it on the SBU \cite{sbu1,sbu2}, ISTD \cite{istd} and UCF \cite{ucf} datasets. We first train our model on SBU dataset, which is the largest publicly available annotated shadow dataset with 4,089 training images and test the model on SBU testing images, ISTD and UCF. We use the Balanced Error Rate (BER) as our evaluation metrics and the same evaluation code as \cite{ber1,sbu1,ber2}. As shown in Table \ref{tab:shadow}, our method outperforms all previous state-of-the-art methods on the benchmark.

\begin{table}[t]
    \renewcommand\arraystretch{5} 
    \scriptsize
    \vspace{-0.1cm}
    \centering
    \tabcolsep=0.2cm
    {
        \begin{tabular}{ c |  c  c  c  c c c c c}
            \toprule
            & scGAN \cite{ber2}& NLDF \cite{nonlocalsalient} & DSS \cite{DSS} & BMPM \cite{wang2018BiDirectional}
            & JDR \cite{istd} & DSC \cite{hu_shadow} & CPD \cite{cascadesodcvpr2019} & \textbf{Ours}
            \\
            \specialrule{0em}{0pt}{1pt}
            \hline
            \specialrule{0em}{1pt}{0pt}
            SBU 
            & 9.10 & 7.02 & 7.00 & 6.17 & 8.14 & 5.59 & 4.19 & {\color{red}\textbf{4.12}}
            \\
            ISTD
            & 8.98 & 7.50  & 10.48 & 7.10 & 7.35 & 8.24 & 6.76 & {\color{red}\textbf{6.16}}
            \\
            UCF 
            & 11.50 & 7.69 & 10.56 & 8.09 & 11.23 & 8.10 & 7.21 & {\color{red}\textbf{6.89}}
            \\            
            \bottomrule
        \end{tabular}
    }
    \caption{Comparisons of models on three shadow detection datasets: SBU, ISTD and UCF. The evaluation metrics is Balanced Error Rate (BER). A smaller BER means a better result.}
    \vspace{-0.2cm}
    \label{tab:shadow}
\end{table}

\begin{table}[t]
    \renewcommand\arraystretch{5} 
    \scriptsize
    \vspace{-0.1cm}
    \centering
    \tabcolsep=0.2cm
    {
        \begin{tabular}{ c |  c c c c c c}
            \toprule
            Methods & 
            PFCN+ \cite{portrait} & 
            NLDF \cite{nonlocalsalient} &
            DSS \cite{DSS} & 
            BMPM \cite{wang2018BiDirectional} & 
            CPD \cite{cascadesodcvpr2019} &
            \textbf{Ours}
            \\
            \specialrule{0em}{0pt}{1pt}
            \hline
            \specialrule{0em}{1pt}{0pt}
            mIoU&
            95.9\% & 
            95.6\% &
            96.2\% & 
            96.2\% & 
            96.6\% &
            {\color{red}\textbf{96.7\%}}
            \\
            \bottomrule
        \end{tabular}
    }
    \caption{Comparisons of models on Portrait Segmentation.}
    \vspace{-0.2cm}
    \label{tab:portrait}
\end{table}

\vspace{-0.2cm}
\subsubsection{Other Frameworks}
\vspace{-0.1cm}
\label{sec:otherframeworks}
Region Refinement Module and Boundary Refinement Loss, as two independent refinement strategies, can be applied to other frameworks generally. Besides the baseline in Table \ref{tab:ablation_bound_layer}, we also use it in Cascaded Partial Decoder (CPD) \cite{cascadesodcvpr2019} (ResNet-50) and R3Net \cite{r3net} (ResNeXt-101), two state-of-the-art SOD frameworks, to verify our designs. In our experiments, we use the default training setting and use fixed initial random seeds for fair comparisons. We simply insert RRM (without the final upsampling stage RRL$_f$) into the two models without any further change.

\vspace{-0.1in}
\paragraph{CPD} We apply two 4-layers RRMs in two aggregation modules of CPD before the final convolution of each aggregation and use the pre-generated boundary masks for BRL during training. As listed in Table \ref{tab:comparison_cpd}, though our reproduced baseline is less effective as claimed in the original paper, RRM and BRL both improve performance of the model on most of the benchmark datasets. They still help the final model (CPD+BRL+RRM) outperform the original one. It is worth noting that each RRM only brings about 3MB increase in terms of the model size. The speeds (on 352$\times$352 input) of CPD and CPD with two RRMs are 47.3 and 36.2 FPS respectively.

\begin{table}[H]
    \renewcommand\arraystretch{5} 
    \scriptsize
    \vspace{-0.2cm}
    \centering
    \tabcolsep=0.23cm
    {
        \begin{tabular}{ c |  c  c  c  c  c c c c c c }
            \toprule
            \multirow{2}{*}{\textit{Method}} & 
            \multicolumn{2}{c}{ECSSD} & 
            \multicolumn{2}{c}{HKU-IS} & 
            \multicolumn{2}{c}{OMRON} &
            \multicolumn{2}{c}{DUTS} &
            \multicolumn{2}{c}{PASCAL-S}\\
             & \textit{F$_{\beta}$}$\uparrow$ & \textit{MAE}$\downarrow$ & \textit{F$_{\beta}$}$\uparrow$ & \textit{MAE}$\downarrow$ & \textit{F$_{\beta}$}$\uparrow$ & \textit{MAE}$\downarrow$ & \textit{F$_{\beta}$}$\uparrow$ & \textit{MAE}$\downarrow$ & \textit{F$_{\beta}$}$\uparrow$ & \textit{MAE}$\downarrow$ \\
            \specialrule{0em}{0pt}{1pt}
            \hline
            \specialrule{0em}{1pt}{0pt}
            
            CPD (original)
            & 0.937 & 0.037 
            & 0.916 & 0.034 
            & 0.826 & 0.056
            & 0.878 & 0.043
            & 0.881 & 0.071  \\ 
            
            CPD (reproduced)
            & 0.938 & 0.039 
            & 0.924 & 0.034 
            & 0.831 & 0.058
            & 0.879 & 0.046 
            & 0.878 & 0.075  \\
            
            CPD + BRL
            & 0.940 & 0.037 
            & 0.927 & 0.031 
            & 0.840 & {\color{red}\textbf{0.056}}
            & 0.882 & 0.044 
            & 0.886 & 0.070 \\
            
            CPD + RRM
            & 0.939 & {\color{red}\textbf{0.036}}
            & 0.927 & 0.032
            & 0.830 & 0.058
            & 0.883 & 0.044
            & 0.881 & 0.070 \\
   
            \textbf{CPD + BRL + RRM}
            & {\color{red}\textbf{0.941}} & {\color{red}\textbf{0.036}}
            & {\color{red}\textbf{0.932}} & {\color{red}\textbf{0.030}}
            & {\color{red}\textbf{0.846}} & 0.059
            & {\color{red}\textbf{0.890}} & {\color{red}\textbf{0.043}}
            & {\color{red}\textbf{0.900}} & {\color{red}\textbf{0.066}} \\         
            
            \bottomrule
            
        \end{tabular}
    }
    \caption{Comparisons of CPD \cite{cascadesodcvpr2019} with/without our proposed RRM and BRL. }
    \vspace{-0.4cm}
    \label{tab:comparison_cpd}
\end{table}

\vspace{-0.2in}
\paragraph{R3Net} On R3Net, we apply a single 4-layers RRM after the high-level integrated
features and before the following Residual Refinement Blocks. R3Net in the original paper is trained on MSRA10K \cite{contrast1} with results refined by CRF\cite{CRF}. We also train the improved models on DUTS-TR/MSRA10K and test them with/without CRF to verify the performance as shown in Table \ref{tab:comparison_r3net_early}. The speeds (on 352$\times$352 input) of R3Net and R3Net+RRM and R3Net+RRM+CRF are 21.8, 19.2 and 2.4 FPS.

\begin{table}[H]
    \renewcommand\arraystretch{5} 
    \scriptsize
    \vspace{-0.2cm}
    \centering
    \tabcolsep=0.2cm
    {
        \begin{tabular}{ c |  c c c c c c c c c c }
            \toprule
            \multirow{2}{*}{\textit{Method}} & 
            \multicolumn{2}{c}{ECSSD} & 
            \multicolumn{2}{c}{HKU-IS} & 
            \multicolumn{2}{c}{OMRON} &
            \multicolumn{2}{c}{DUTS} &
            \multicolumn{2}{c}{PASCAL-S}\\
             & \textit{F$_{\beta}$}$\uparrow$ & \textit{MAE}$\downarrow$ & \textit{F$_{\beta}$}$\uparrow$ & \textit{MAE}$\downarrow$ & \textit{F$_{\beta}$}$\uparrow$ & \textit{MAE}$\downarrow$ & \textit{F$_{\beta}$}$\uparrow$ & \textit{MAE}$\downarrow$ & \textit{F$_{\beta}$}$\uparrow$ & \textit{MAE}$\downarrow$ \\

            \specialrule{0em}{0pt}{1pt}
            \hline
            \specialrule{0em}{1pt}{0pt}
            \multicolumn{11}{c}{DUTS + Non-CRF} \\
            \specialrule{0em}{0pt}{1pt}
            \hline
            \specialrule{0em}{1pt}{0pt}    
            R3Net-D 
            & 0.943 & 0.042
            & 0.932 & 0.038 
            & 0.840 & 0.065 
            & 0.896 & 0.045 
            & 0.895 & 0.072  \\  
            
            R3Net-D + BRL
            & 0.946 & 0.037 
            & {\color{red}\textbf{0.936}} & 0.035 
            & 0.844 & 0.063 
            & 0.901 & {\color{red}\textbf{0.040}} 
            & 0.897 & 0.068  \\      
            
            R3Net-D + RRM
            & 0.943 & 0.040 
            & 0.933 & 0.034 
            & 0.843 & 0.063
            & 0.900 & 0.042 
            & {\color{red}\textbf{0.899}} & 0.069  \\ 
            
            \textbf{R3Net-D + RRM + BRL}
            & {\color{red}\textbf{0.947}} & {\color{red}\textbf{0.035}}
            & {\color{red}\textbf{0.936}} & {\color{red}\textbf{0.030}}
            & {\color{red}\textbf{0.850}} & {\color{red}\textbf{0.062}}
            & {\color{red}\textbf{0.903}} & {\color{red}\textbf{0.040}}
            & {\color{red}\textbf{0.899}} & {\color{red}\textbf{0.066}}  \\

            \specialrule{0em}{0pt}{1pt}
            \hline
            \specialrule{0em}{1pt}{0pt}
            \multicolumn{11}{c}{MSRA10K + CRF} \\
            \specialrule{0em}{0pt}{1pt}
            \hline
            \specialrule{0em}{1pt}{0pt}
            R3Net-M (original)
            & 0.934 & 0.040 
            & 0.922 & 0.036 
            & 0.841 & 0.063 
            & 0.866 & 0.057 
            & 0.841 & 0.092  \\  
            
            R3Net-M (reproduced)
            & 0.936 & 0.040 
            & 0.927 & 0.034 
            & 0.845 & 0.069 
            & 0.866 & 0.062 
            & 0.843 & 0.092  \\  
            
            R3Net-M + BRL
            & 0.938 & 0.039 
            & 0.928 & 0.034 
            & 0.845 & {\color{red}\textbf{0.065}} 
            & 0.870 & {\color{red}\textbf{0.060}} 
            & 0.844 & 0.089  \\ 
            
            R3Net-M + RRM
            & 0.936 & 0.038 
            & 0.929 & 0.032 
            & 0.846 & 0.067 
            & 0.869 & {\color{red}\textbf{0.060}}
            & 0.853 & {\color{red}\textbf{0.087}}  \\             
            
            \textbf{R3Net-M + RRM + BRL}
            & {\color{red}\textbf{0.940}} & {\color{red}\textbf{0.038}} 
            & {\color{red}\textbf{0.931}} & {\color{red}\textbf{0.031}}
            & {\color{red}\textbf{0.851}} & 0.068 
            & {\color{red}\textbf{0.875}} & {\color{red}\textbf{0.060}} 
            & {\color{red}\textbf{0.857}} & 0.088  \\

            \bottomrule
        \end{tabular}
    }
    \caption{Comparisons of R3Net with/without RRM and BRL. R3Net-D: Model trained on DUTS training data. R3Net-M: Model trained on MSRA10K. `Original': the performance reported in paper.}
    \vspace{-0.5cm}
    \label{tab:comparison_r3net_early}
\end{table}

\vspace{-0.4cm}
\section{Conclusion}
\vspace{-0.1in}
We have presented a novel Region Refinement Module (RRM) to optimize saliency prediction with a gating mechanism. Our boundary refinement loss (BRL) refines the boundary directly on the prediction without introducing new parameters and extra computation. These two designs not only help our Region Refinement Network (RRN) achieve new state-of-the-art results on five popular benchmark datasets but also can be easily applied to other frameworks. Possible future work includes extending these designs to video saliency detection and further improving the performance.



\clearpage
\vspace{-0.3cm}
\section{Appendix}
\vspace{-0.3cm}
This section is the supplementary material for Region Refinement Network for Salient Object
Detection. In Section \ref{sec:error_bar}, we illustrate the mean, variance and error bars of our model. More experiments on our proposed Region Refinement Module (RRM) and Boundary Refinement Loss (BRL) can be found in Section \ref{sec:more_detail}. In Section \ref{sec:more_exp} and Section \ref{sec:links}, we show more examples for visual comparison and links for publicly available datasets used in our experiments.

\vspace{-0.3cm}
\subsection{Mean, Variance and Error Bar}
\label{sec:error_bar}
\begin{figure}[H]
\vspace{-0.1cm}
\centering
    \begin{minipage} [t] {0.35\linewidth}
        \centering
        \includegraphics [width=1.0\linewidth,height=0.75\linewidth] {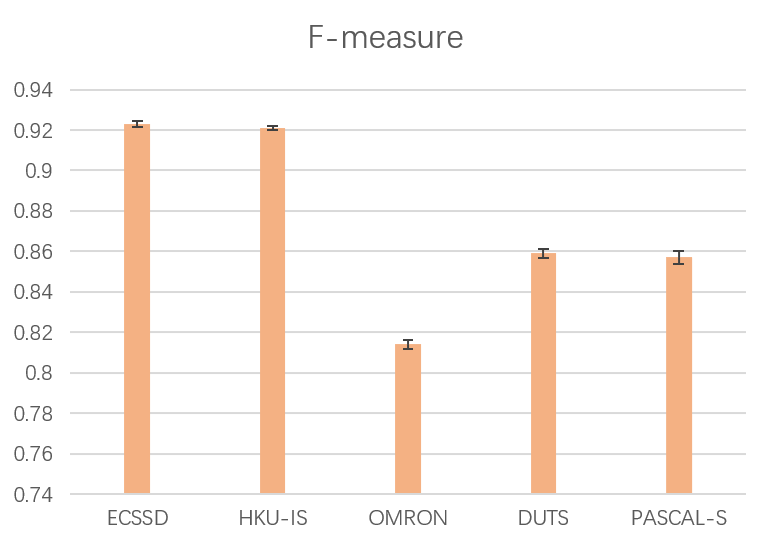}
        {(a)}
    \end{minipage} 
    \begin{minipage} [t] {0.35\linewidth}
        \centering
        \includegraphics [width=1.0\linewidth,height=0.75\linewidth] {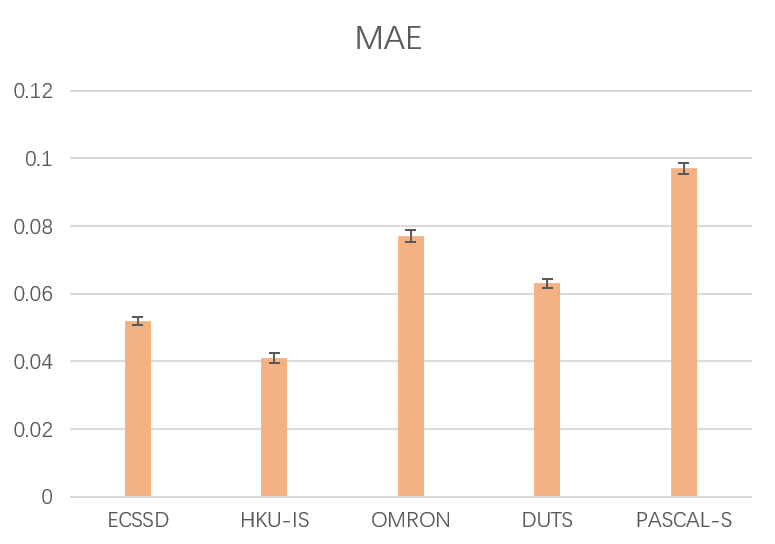}
        {(b)}
    \end{minipage}
    \caption{The error bar of F-measure (a) and MAE (b) on five datasets.}
    \label{fig:errorbar}
    \vspace{-0.2cm}
\end{figure}

\vspace{-0.1cm}
We train extra 10 models to show the mean, variance and error bars. Due to the limited computing resources, extra experiments are based on VGG16 backbone and trained 50 epochs on MSRA-B (2500 images) dataset to have a quick analysis. Finally, we evaluate these models on HKU-IS \cite{HKU-IS} (4447 images), 
ECSSD \cite{ECSSD} (1000 images) 
DUTS \cite{DUTS} (5018 images), 
DUT-OMRON \cite{OMRON} (5168 images) 
and PASCAL-S \cite{PASCAL-S} (850 images). The results are shown in Table \ref{tab:errorbar} and Figure \ref{fig:errorbar}.

\begin{table}[H]
    \renewcommand\arraystretch{5} 
    \scriptsize
    \vspace{-0.2cm}
    \centering
    \tabcolsep=0.2cm
    {
        \begin{tabular}{ c |  c c c c c c c c c c }
            \toprule
            \multirow{2}{*}{\textit{Model}} & 
            \multicolumn{2}{c}{ECSSD} & 
            \multicolumn{2}{c}{HKU-IS} & 
            \multicolumn{2}{c}{OMRON} &
            \multicolumn{2}{c}{DUTS} &
            \multicolumn{2}{c}{PASCAL-S}\\
             & \textit{F$_{\beta}$}$\uparrow$ & \textit{MAE}$\downarrow$ & \textit{F$_{\beta}$}$\uparrow$ & \textit{MAE}$\downarrow$ & \textit{F$_{\beta}$}$\uparrow$ & \textit{MAE}$\downarrow$ & \textit{F$_{\beta}$}$\uparrow$ & \textit{MAE}$\downarrow$ & \textit{F$_{\beta}$}$\uparrow$ & \textit{MAE}$\downarrow$ \\
            \specialrule{0em}{0pt}{1pt}
            \hline
            \specialrule{0em}{1pt}{0pt}
            
            1
            & 0.925 & 0.054 
            & 0.921 & 0.043 
            & 0.813 & 0.079 
            & 0.858 & 0.064 
            & 0.863 & 0.096  \\  
            
            2
            & 0.924 & 0.050 
            & 0.921 & 0.040 
            & 0.813 & 0.075 
            & 0.861 & 0.061 
            & 0.853 & 0.098  \\  
            
            3
            & 0.921 & 0.052 
            & 0.922 & 0.042 
            & 0.814 & 0.075 
            & 0.861 & 0.062 
            & 0.856 & 0.096  \\  
            
            4
            & 0.922 & 0.051 
            & 0.921 & 0.040 
            & 0.815 & 0.079 
            & 0.861 & 0.063 
            & 0.855 & 0.096  \\

            5
            & 0.922 & 0.052 
            & 0.919 & 0.041 
            & 0.820 & 0.077 
            & 0.857 & 0.065 
            & 0.855 & 0.098  \\  
            
            6
            & 0.921 & 0.053 
            & 0.922 & 0.044 
            & 0.811 & 0.075 
            & 0.859 & 0.062 
            & 0.854 & 0.096  \\   
            
            7
            & 0.921 & 0.051
            & 0.920 & 0.040 
            & 0.816 & 0.077 
            & 0.862 & 0.062 
            & 0.856 & 0.096  \\  
            
            8
            & 0.923 & 0.051
            & 0.922 & 0.041 
            & 0.815 & 0.078 
            & 0.860 & 0.064 
            & 0.862 & 0.095  \\  
            
            9
            & 0.924 & 0.052 
            & 0.922 & 0.039 
            & 0.814 & 0.075 
            & 0.861 & 0.062 
            & 0.856 & 0.100  \\  
            
            10
            & 0.925 & 0.050
            & 0.921 & 0.040 
            & 0.813 & 0.079 
            & 0.855 & 0.064 
            & 0.860 & 0.094  \\  
            
            \specialrule{0em}{0pt}{1pt}
            \hline
            \specialrule{0em}{1pt}{0pt}
            \textbf{Mean}
            & 0.923 & 0.052
            & 0.921 & 0.041 
            & 0.814 & 0.077 
            & 0.859 & 0.063
            & 0.857 & 0.097  \\ 
            
            \textbf{Std ($\times10^{-3}$)}
            & 1.536 & 1.200
            & 0.943 & 1.483 
            & 2.289 & 1.700
            & 2.110 & 1.221
            & 3.256 & 1.628  \\

            \bottomrule
        \end{tabular}
    }
    \caption{Extra 10 models to show the mean and standard deviation.}
    \vspace{-0.3cm}
    \label{tab:errorbar}
\end{table}

\begin{table}[H]
    \renewcommand\arraystretch{5} 
    \scriptsize
    \centering
    \tabcolsep=0.2cm
    {
        \begin{tabular}{ c |  c c c c c c c c c c }
            \toprule
            \multirow{2}{*}{\textit{Method}} & 
            \multicolumn{2}{c}{ECSSD} & 
            \multicolumn{2}{c}{HKU-IS} & 
            \multicolumn{2}{c}{OMRON} &
            \multicolumn{2}{c}{DUTS} &
            \multicolumn{2}{c}{PASCAL-S}\\
             & \textit{F$_{\beta}$}$\uparrow$ & \textit{MAE}$\downarrow$ & \textit{F$_{\beta}$}$\uparrow$ & \textit{MAE}$\downarrow$ & \textit{F$_{\beta}$}$\uparrow$ & \textit{MAE}$\downarrow$ & \textit{F$_{\beta}$}$\uparrow$ & \textit{MAE}$\downarrow$ & \textit{F$_{\beta}$}$\uparrow$ & \textit{MAE}$\downarrow$ \\
              \specialrule{0em}{0pt}{1pt}
            \hline
            \specialrule{0em}{1pt}{0pt}
            Baseline
            & 0.936 & 0.043 & 0.931 & 0.035 & 0.819 & 0.063 & 0.886 & 0.044 & 0.878 & 0.079  \\
            Baseline+BRL
            & 0.943 & 0.038 & 0.934 & 0.030 & 0.833 & 0.061 & 0.889 & 0.042 & 0.882 & 0.076  \\
            Baseline+BRL+RRM$_{1}$
            & 0.945 & 0.035 & 0.938 & 0.028 & {\color{red}\textbf{0.835}} & 0.059 & 0.892 & 0.040 & 0.883 & 0.074  \\
            Baseline+BRL+RRM$_{2}$
            & 0.942 & 0.035 & 0.939 & {\color{red}\textbf{0.027}} & 0.817 & 0.056 & 0.892 & {\color{red}\textbf{0.038}} & 0.882 & 0.072  \\
            Baseline+BRL+RRM$_{3}$
            & 0.946 & 0.035 & {\color{red}\textbf{0.941}} & {\color{red}\textbf{0.027}} & {\color{red}\textbf{0.835}} & 0.055 & 0.893 & {\color{red}\textbf{0.038}} & 0.883 & 0.072  \\
            
            \textbf{Baseline+BRL+RRM$_{4}$}
            & 0.946 & {\color{red}\textbf{0.034}} & {\color{red}\textbf{0.941}} & {\color{red}\textbf{0.027}} & 0.833 & {\color{red}\textbf{0.054}} & 0.893 & {\color{red}\textbf{0.038}} & {\color{red}\textbf{0.900}} & {\color{red}\textbf{0.067}}  \\
            
            Baseline+BRL+RRM$_{5}$
            & 0.946 & 0.035 & 0.940 & {\color{red}\textbf{0.027}} & 0.833 & 0.056 & 0.884 & 0.039 & 0.894 & 0.070 \\
            Baseline+BRL+RRM$_{6}$
            & {\color{red}\textbf{0.947}} & 0.034 & 0.938 & {\color{red}\textbf{0.027}} & {\color{red}\textbf{0.835}} & 0.057 & {\color{red}\textbf{0.895}} & 0.039 & 0.892 & 0.071  \\
            Baseline+BRL+RRM$_{7}$
            & 0.945 & 0.035 & 0.839 & 0.028 & 0.832 & 0.057 & 0.888 & 0.040 & 0.888 & 0.070  \\
            
            \bottomrule
        \end{tabular}
    }
    \vspace{0.1cm}
    \caption{Comparisons of different layer numbers $N$. All models use ResNet50 as the backbone and are trained on DUTS training data.}
    \vspace{-0.5cm}
    \label{tab:ablation_bound_layer}
\end{table}

\subsection{More Details and Ablation Studies}
\label{sec:more_detail}

\subsubsection{Region Refinement Module (RRM)}
In the proposed Regional Refinement Network (RRN) and applications on CPD\cite{cascadesodcvpr2019} and R3Net\cite{r3net}, the input features have 64, 96 and 256 output channels respectively (we process the input feature in RRN by a 1$\times$1 convolution with 64 output channels), therefore all intermediate 3$\times$3 and 1$\times$1 convolutions in RRM have same input and output channels, except for the last convolution in each RRL (Region Refinement Layer) outputs a salient map with 1 output channel. 

\paragraph{Layer Number ($N$)}
The results of different N are shown in Table \ref{tab:ablation_bound_layer}. Note that when $N$ is set to zero, the model degrades to Baseline+BRL in Table \ref{tab:ablation_bound_layer}. Our results from RRM$_{0}$ to RRM$_{7}$ are gradually refined, decent in terms of both $F_{\beta}$ and $MAE$. We observe that the model with RRM$_{4}$ is optimal in most cases even if it contains fewer parameters than RRM$_{N} (N > 4)$. We thus choose to apply RRM$_{4}$ in our final model.

\paragraph{Number of 3$\times$3 and 1$\times$1 Convolution ($M$)}
Then, apart from the vertical analysis on layer number $N$, we further perform horizontal analysis on the number of convolutions. Since the numbers of 1$\times$1 convolutions and 3$\times$3 convolutions are equal in each Region Refinement Layer (RRL), we use $M$ to denote the numbers of two convolutions in each RRL. The results of $M$ on RRN are as Table \ref{tab:convlayers} shows. $M = 3$ yields the best performance.

\begin{table}[H]
    \renewcommand\arraystretch{5} 
    \scriptsize
    \centering
    \tabcolsep=0.2cm
    {
        \begin{tabular}{ c |  c c c c c c c c c c }
            \toprule
            \multirow{2}{*}{\textit{M}} & 
            \multicolumn{2}{c}{ECSSD} & 
            \multicolumn{2}{c}{HKU-IS} & 
            \multicolumn{2}{c}{OMRON} &
            \multicolumn{2}{c}{DUTS} &
            \multicolumn{2}{c}{PASCAL-S}\\
             & \textit{F$_{\beta}$}$\uparrow$ & \textit{MAE}$\downarrow$ & \textit{F$_{\beta}$}$\uparrow$ & \textit{MAE}$\downarrow$ & \textit{F$_{\beta}$}$\uparrow$ & \textit{MAE}$\downarrow$ & \textit{F$_{\beta}$}$\uparrow$ & \textit{MAE}$\downarrow$ & \textit{F$_{\beta}$}$\uparrow$ & \textit{MAE}$\downarrow$ \\
            \specialrule{0em}{0pt}{1pt}
            \hline
            \specialrule{0em}{1pt}{0pt}
            0
            & 0.925 & 0.048
            & 0.928 & 0.036 
            & 0.795 & 0.074 
            & 0.880 & 0.048 
            & 0.881 & 0.081  \\            
            
            1
            & 0.930 & 0.044 
            & 0.929 & 0.033 
            & 0.804 & 0.067 
            & 0.885 & 0.042 
            & 0.879 & 0.075  \\
            
            2
            & 0.925 & 0.041 
            & 0.930 & {\color{red}\textbf{0.031}} 
            & 0.797 & 0.062 
            & 0.885 & 0.041 
            & 0.883 & {\color{red}\textbf{0.072}}  \\
            
            3
            & {\color{red}\textbf{0.936}} & {\color{red}\textbf{0.040}} 
            & {\color{red}\textbf{0.932}} & {\color{red}\textbf{0.031}} 
            & 0.809 & {\color{red}\textbf{0.060}}
            & 0.884 & {\color{red}\textbf{0.040}} 
            & {\color{red}\textbf{0.886}} & {\color{red}\textbf{0.072}}  \\
            
            4
            & 0.933 & 0.042 
            & 0.930 & {\color{red}\textbf{0.031}} 
            & {\color{red}\textbf{0.811}} & 0.063 
            & {\color{red}\textbf{0.890}} & 0.042 
            & 0.883 & 0.073  \\

            \bottomrule
        \end{tabular}
    }
    \vspace{0.1cm}
    \caption{Comparisons of different numbers of 3$\times$3 and 1$\times$1 convolutions in each RRL. All models ($N=4$) are based on VGG16 and trained on DUTS training data\cite{DUTS}. $M=0$ means no RRM.}
    \vspace{0cm}
    \label{tab:convlayers}
\end{table}

\begin{figure}[H]
    \vspace{-0.1cm}
    \centering
    \begin{minipage} [t] {0.1\linewidth}
        \centering
        \includegraphics [width=1\linewidth,height=0.6\linewidth] 
        {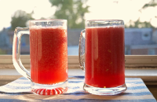}
    \end{minipage}
    \begin{minipage} [t] {0.1\linewidth}
        \centering
        \includegraphics [width=1\linewidth,height=0.75\linewidth] {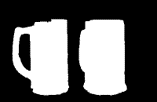}
    \end{minipage}
    \begin{minipage} [t] {0.1\linewidth}
        \centering
        \includegraphics [width=1\linewidth,height=0.75\linewidth] {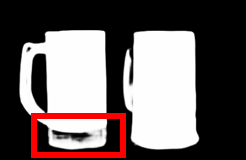}
     \end{minipage}
    \begin{minipage} [t] {0.1\linewidth}
        \centering
        \includegraphics [width=1\linewidth,height=0.75\linewidth] {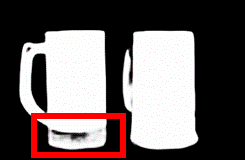}
    \end{minipage}
    \begin{minipage} [t] {0.1\linewidth}
        \centering
        \includegraphics [width=1\linewidth,height=0.75\linewidth] {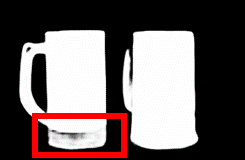}
    \end{minipage}    
    \begin{minipage} [t] {0.1\linewidth}
        \centering
        \includegraphics [width=1\linewidth,height=0.75\linewidth] {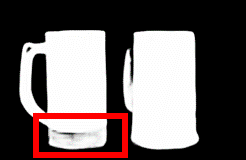}
    \end{minipage}     

    \begin{minipage} [t] {0.1\linewidth}
        \centering
        \includegraphics [width=1\linewidth,height=0.75\linewidth] 
        {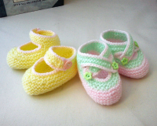}
    \end{minipage}
    \begin{minipage} [t] {0.1\linewidth}
        \centering
        \includegraphics [width=1\linewidth,height=0.75\linewidth] {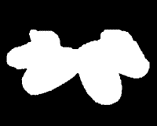}
    \end{minipage}
    \begin{minipage} [t] {0.1\linewidth}
        \centering
        \includegraphics [width=1\linewidth,height=0.75\linewidth] {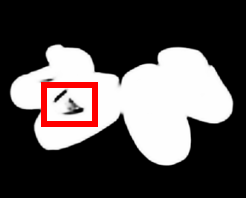}
     \end{minipage}
    \begin{minipage} [t] {0.1\linewidth}
        \centering
        \includegraphics [width=1\linewidth,height=0.75\linewidth] {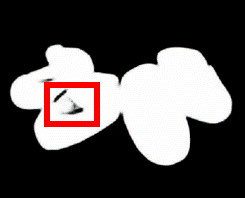}
    \end{minipage}
    \begin{minipage} [t] {0.1\linewidth}
        \centering
        \includegraphics [width=1\linewidth,height=0.75\linewidth] {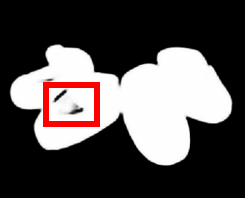}
    \end{minipage}    
    \begin{minipage} [t] {0.1\linewidth}
        \centering
        \includegraphics [width=1\linewidth,height=0.75\linewidth] {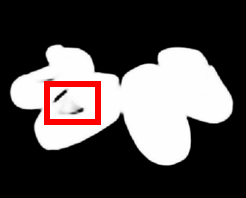}
    \end{minipage}     

    \begin{minipage} [t] {0.1\linewidth}
        \centering
        \includegraphics [width=1\linewidth,height=0.75\linewidth] 
        {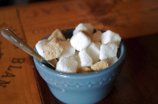}
    \end{minipage}
    \begin{minipage} [t] {0.1\linewidth}
        \centering
        \includegraphics [width=1\linewidth,height=0.75\linewidth] {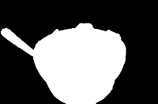}
    \end{minipage}
    \begin{minipage} [t] {0.1\linewidth}
        \centering
        \includegraphics [width=1\linewidth,height=0.75\linewidth] {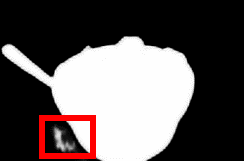}
     \end{minipage}
    \begin{minipage} [t] {0.1\linewidth}
        \centering
        \includegraphics [width=1\linewidth,height=0.75\linewidth] {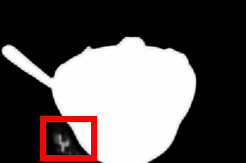}
    \end{minipage}
    \begin{minipage} [t] {0.1\linewidth}
        \centering
        \includegraphics [width=1\linewidth,height=0.75\linewidth] {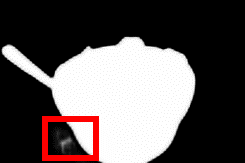}
    \end{minipage}    
    \begin{minipage} [t] {0.1\linewidth}
        \centering
        \includegraphics [width=1\linewidth,height=0.75\linewidth] {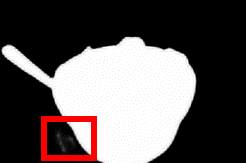}
    \end{minipage}         

    \begin{minipage} [t] {0.1\linewidth}
        \centering
        \includegraphics [width=1\linewidth,height=0.75\linewidth] 
        {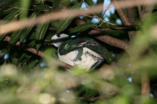}
        {(a)}
    \end{minipage}
    \begin{minipage} [t] {0.1\linewidth}
        \centering
        \includegraphics [width=1\linewidth,height=0.75\linewidth] {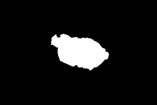}
        {(b)}
    \end{minipage}
    \begin{minipage} [t] {0.1\linewidth}
        \centering
        \includegraphics [width=1\linewidth,height=0.75\linewidth] {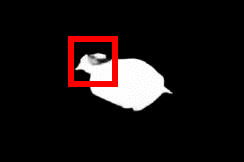}
        {(c)}
     \end{minipage}
    \begin{minipage} [t] {0.1\linewidth}
        \centering
        \includegraphics [width=1\linewidth,height=0.75\linewidth] {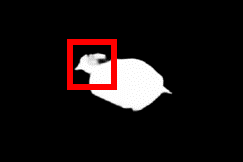}
        {(d)}
    \end{minipage}
    \begin{minipage} [t] {0.1\linewidth}
        \centering
        \includegraphics [width=1\linewidth,height=0.75\linewidth] {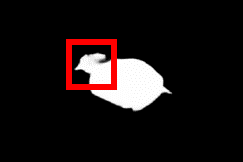}
        {(e)}
    \end{minipage}    
    \begin{minipage} [t] {0.1\linewidth}
        \centering
        \includegraphics [width=1\linewidth,height=0.75\linewidth] {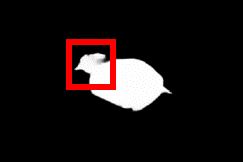}
        {(f)}
    \end{minipage}         
    \vspace{-0.1cm}
    \caption{Visualization of the refinement process. (a) Input image. (b) Ground truth. (c)-(f) are outputs of $RRL_{1}$, $RRL_{2}$, $RRL_{3}$ and $RRL_{4}$. }
    \label{fig:vis_refine}
    \vspace{-0.2cm}
\end{figure}

\paragraph{RRM Refinement Process}
We take outputs from $RRL_{1}$, $RRL_{2}$, $RRL_{3}$ and $RRL_{4}$ to better illustrate the refining process in RRM. As shown in Figure \ref{fig:vis_refine}, false predictions are gradually eliminated from $RRL_{1}$ to $RRL_{4}$.

\subsubsection{Boundary Refinement Learning (BRL)}
\paragraph{Boundary Mask Generation}
In BRL, we use boundary mask for extra supervision on boundary area. In this section, we show the impacts of different boundary widths. As shown in Table \ref{tab:widths}, $Width=5$ yields the best performance.

\begin{table}[H]
    \renewcommand\arraystretch{5} 
    \scriptsize
    \vspace{-0.2cm}
    \centering
    \tabcolsep=0.2cm
    {
        \begin{tabular}{ c |  c c c c c c c c c c }
            \toprule
            \multirow{2}{*}{\textit{Method}} & 
            \multicolumn{2}{c}{ECSSD} & 
            \multicolumn{2}{c}{HKU-IS} & 
            \multicolumn{2}{c}{OMRON} &
            \multicolumn{2}{c}{DUTS} &
            \multicolumn{2}{c}{PASCAL-S}\\
             & \textit{F$_{\beta}$}$\uparrow$ & \textit{MAE}$\downarrow$ & \textit{F$_{\beta}$}$\uparrow$ & \textit{MAE}$\downarrow$ & \textit{F$_{\beta}$}$\uparrow$ & \textit{MAE}$\downarrow$ & \textit{F$_{\beta}$}$\uparrow$ & \textit{MAE}$\downarrow$ & \textit{F$_{\beta}$}$\uparrow$ & \textit{MAE}$\downarrow$ \\
             
            \specialrule{0em}{0pt}{1pt}
            \hline
            \specialrule{0em}{1pt}{0pt}
            No Bound
            & 0.917 & 0.061 
            & 0.915 & 0.050
            & 0.804 & 0.088 
            & 0.852 & 0.074 
            & 0.853 & 0.105  \\   
            
            \textit{Width}=0
            & 0.917 & 0.056 
            & 0.918 & 0.048 
            & 0.806 & 0.079 
            & 0.853 & 0.073 
            & 0.849 & 0.099  \\            
            
            \textit{Width}=1
            & 0.923 & 0.052 
            & 0.918 & 0.045
            & 0.812 & 0.086 
            & 0.855 & 0.071 
            & 0.855 & 0.100  \\ 
            
            \textit{Width}=3
            & 0.923 & 0.051 
            & {\color{red}\textbf{0.921}} & 0.042 
            & 0.808 & 0.085 
            & 0.855 & 0.068 
            & {\color{red}\textbf{0.862}} & 0.099  \\ 
            
            \textit{Width}=5
            & {\color{red}\textbf{0.925}} & {\color{red}\textbf{0.050}}
            & {\color{red}\textbf{0.921}} & {\color{red}\textbf{0.039}}
            & 0.810 & 0.073 
            & {\color{red}\textbf{0.861}} & {\color{red}\textbf{0.058}} 
            & 0.852 & {\color{red}\textbf{0.096}}  \\ 
            
            \textit{Width}=7
            & 0.924 & {\color{red}\textbf{0.050}} 
            & 0.920 & 0.041 
            & 0.807 & 0.077 
            & 0.857 & 0.062 
            & 0.855 & 0.098  \\ 

            \textit{Width}=9
            & 0.922 & 0.051 
            & 0.918 & 0.040 
            & 0.809 & {\color{red}\textbf{0.072}} 
            & 0.860 & {\color{red}\textbf{0.058}} 
            & 0.851 & 0.098  \\ 
            
            \textit{Width}=11
            & 0.921 & 0.054 
            & 0.920 & 0.045 
            & {\color{red}\textbf{0.814}} & 0.090 
            & 0.853 & 0.074 
            & 0.854 & 0.102  \\            
            \bottomrule
        \end{tabular}
    }
    \caption{Comparisons of different boundary widths in BRL. All models ($N=4, M=3$) are based on VGG16 and trained on MSRA-B training data\cite{DUTS}. `No Bound' means BRL is not implemented. $Width=0$ means the `Grad' enhancement strategy of \cite{nonlocalsalient} is implemented.}
    \vspace{-0.3cm}
    \label{tab:widths}
\end{table}

\vspace{-0.3cm}
\subsubsection{More Results of R3Net}
\vspace{-0.1cm}
We provide the detailed results of our application on R3Net \cite{r3net} in Table \ref{tab:comparison_r3net} where we add the performance of DUTS+CRF and MSRA10K+Non-CRF to the original table.

\begin{table}[H]
    \renewcommand\arraystretch{5} 
    \scriptsize
    \vspace{-0.2cm}
    \centering
    \tabcolsep=0.2cm
    {
        \begin{tabular}{ c |  c c c c c c c c c c }
            \toprule
            \multirow{2}{*}{\textit{Method}} & 
            \multicolumn{2}{c}{ECSSD} & 
            \multicolumn{2}{c}{HKU-IS} & 
            \multicolumn{2}{c}{OMRON} &
            \multicolumn{2}{c}{DUTS} &
            \multicolumn{2}{c}{PASCAL-S}\\
             & \textit{F$_{\beta}$}$\uparrow$ & \textit{MAE}$\downarrow$ & \textit{F$_{\beta}$}$\uparrow$ & \textit{MAE}$\downarrow$ & \textit{F$_{\beta}$}$\uparrow$ & \textit{MAE}$\downarrow$ & \textit{F$_{\beta}$}$\uparrow$ & \textit{MAE}$\downarrow$ & \textit{F$_{\beta}$}$\uparrow$ & \textit{MAE}$\downarrow$ \\

            \specialrule{0em}{0pt}{1pt}
            \hline
            \specialrule{0em}{1pt}{0pt}
            \multicolumn{11}{c}{DUTS+Non-CRF} \\
            \specialrule{0em}{0pt}{1pt}
            \hline
            \specialrule{0em}{1pt}{0pt}   
            R3Net-D 
            & 0.943 & 0.042
            & 0.932 & 0.038 
            & 0.840 & 0.065 
            & 0.896 & 0.045 
            & 0.895 & 0.072  \\  
            
            R3Net-D + BRL
            & 0.946 & 0.037 
            & {\color{red}\textbf{0.936}} & 0.035 
            & 0.844 & 0.063 
            & 0.901 & {\color{red}\textbf{0.040}} 
            & 0.897 & 0.068  \\      
            
            R3Net-D + RRM
            & 0.943 & 0.040 
            & 0.933 & 0.034 
            & 0.843 & 0.063
            & 0.900 & 0.042 
            & {\color{red}\textbf{0.899}} & 0.069  \\ 
            
            \textbf{R3Net-D + RRM + BRL}
            & {\color{red}\textbf{0.947}} & {\color{red}\textbf{0.035}}
            & {\color{red}\textbf{0.936}} & {\color{red}\textbf{0.030}}
            & {\color{red}\textbf{0.850}} & {\color{red}\textbf{0.062}}
            & {\color{red}\textbf{0.903}} & {\color{red}\textbf{0.040}}
            & {\color{red}\textbf{0.899}} & {\color{red}\textbf{0.066}}  \\ 
            
            \specialrule{0em}{0pt}{1pt}
            \hline
            \specialrule{0em}{1pt}{0pt}
            \multicolumn{11}{c}{DUTS + CRF} \\       
            \specialrule{0em}{0pt}{1pt}
            \hline
            \specialrule{0em}{1pt}{0pt}
            R3Net-D 
            & 0.945 & 0.032 
            & 0.936 & 0.028 
            & 0.840 & 0.056 
            & 0.900 & 0.038 
            & 0.893 & 0.065  \\  
            
            R3Net-D + BRL
            & 0.947 & 0.032 
            & {\color{red}\textbf{0.938}} & 0.027 
            & 0.843 & 0.058 
            & 0.904 & 0.036 
            & 0.898 & 0.064  \\    
            
            R3Net-D + RRM
            & 0.946 & 0.032 
            & 0.936 & 0.027 
            & 0.845 & {\color{red}\textbf{0.055}} 
            & 0.901 & {\color{red}\textbf{0.035}} 
            & {\color{red}\textbf{0.899}} & 0.062  \\    
            
            \textbf{R3Net-D + RRM + BRL}
            & {\color{red}\textbf{0.949}} &{\color{red}\textbf{0.030}} 
            & {\color{red}\textbf{0.938}} & {\color{red}\textbf{0.026}} 
            & {\color{red}\textbf{0.849}} & 0.058 
            & {\color{red}\textbf{0.906}} & 0.036 
            & {\color{red}\textbf{0.899}} & {\color{red}\textbf{0.061}}  \\              

            \specialrule{0em}{0pt}{1pt}
            \hline
            \specialrule{0em}{1pt}{0pt}
            \multicolumn{11}{c}{MSRA10K + Non-CRF} \\
            \specialrule{0em}{0pt}{1pt}
            \hline
            \specialrule{0em}{1pt}{0pt}           
            R3Net-M 
            & 0.933 & 0.050 
            & 0.921 & 0.046 
            & 0.841 & 0.078 
            & 0.862 & 0.070 
            & 0.852 & 0.100  \\     
            
            R3Net-M + BRL
            & 0.936 & 0.044
            & 0.922 & 0.040 
            & 0.842 & {\color{red}\textbf{0.072}} 
            & 0.866 & 0.067 
            & 0.857 & 0.097  \\     
            
            R3Net-M + RRM 
            & 0.938 & 0.047 
            & 0.924 & 0.044 
            & 0.845 & 0.077 
            & 0.868 & 0.066 
            & 0.859 & 0.095  \\     
            
            \textbf{R3Net-M + RRM + BRL}
            & {\color{red}\textbf{0.939}} & {\color{red}\textbf{0.042}}
            & {\color{red}\textbf{0.928}} & {\color{red}\textbf{0.037}} 
            & {\color{red}\textbf{0.848}} & 0.074 
            & {\color{red}\textbf{0.871}} & {\color{red}\textbf{0.065}} 
            & {\color{red}\textbf{0.866}} & {\color{red}\textbf{0.091}}  \\                 
            
            \specialrule{0em}{0pt}{1pt}
            \hline
            \specialrule{0em}{1pt}{0pt}
            \multicolumn{11}{c}{MSRA10K + CRF} \\
            \specialrule{0em}{0pt}{1pt}
            \hline
            \specialrule{0em}{1pt}{0pt}
            R3Net-M (original)
            & 0.934 & 0.040 
            & 0.922 & 0.036 
            & 0.841 & 0.063 
            & 0.866 & 0.057 
            & 0.841 & 0.092  \\  
            
            R3Net-M (reproduced)
            & 0.936 & 0.040 
            & 0.927 & 0.034 
            & 0.845 & 0.069 
            & 0.866 & 0.062 
            & 0.843 & 0.092  \\  
            
            R3Net-M + BRL
            & 0.938 & 0.039 
            & 0.928 & 0.034 
            & 0.845 & {\color{red}\textbf{0.065}} 
            & 0.870 & {\color{red}\textbf{0.060}} 
            & 0.844 & 0.089  \\ 
            
            R3Net-M + RRM
            & 0.936 & 0.038 
            & 0.929 & 0.032 
            & 0.846 & 0.067 
            & 0.869 & {\color{red}\textbf{0.060}}
            & 0.853 & {\color{red}\textbf{0.087}}  \\             
            
            \textbf{R3Net-M + RRM + BRL}
            & {\color{red}\textbf{0.940}} & {\color{red}\textbf{0.038}} 
            & {\color{red}\textbf{0.931}} & {\color{red}\textbf{0.031}}
            & {\color{red}\textbf{0.851}} & 0.068 
            & {\color{red}\textbf{0.875}} & {\color{red}\textbf{0.060}} 
            & {\color{red}\textbf{0.857}} & 0.088  \\    

            \bottomrule
        \end{tabular}
    }
    \caption{Comparisons of R3Net with/without RRM and BRL. R3Net-D: Model trained on DUTS training data. R3Net-M: Model trained on MSRA10K. `Original': the performance reported in paper.}
    \vspace{-0.5cm}
    \label{tab:comparison_r3net}
\end{table}

\vspace{-0.2cm}
\subsection{More Examples of Visual Comparisons}
\label{sec:more_exp}
\begin{figure}[H]
    \begin{minipage} [t] {0.095\linewidth}
        \centering
        \includegraphics [width=1\linewidth,height=0.75\linewidth] {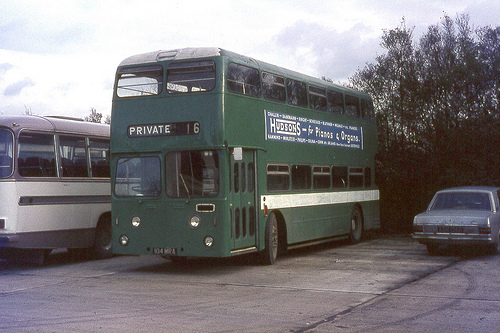}
    \end{minipage}
    \begin{minipage} [t] {0.095\linewidth}
        \centering
        \includegraphics [width=1\linewidth,height=0.75\linewidth] {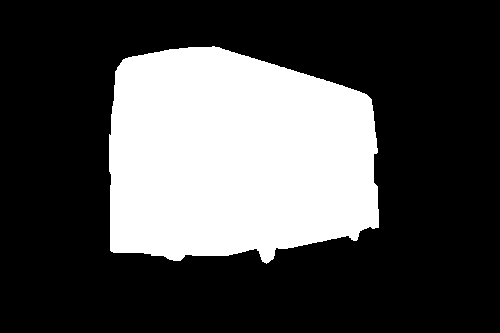}
    \end{minipage}
    \begin{minipage} [t] {0.095\linewidth}
        \centering
        \includegraphics [width=1\linewidth,height=0.75\linewidth] {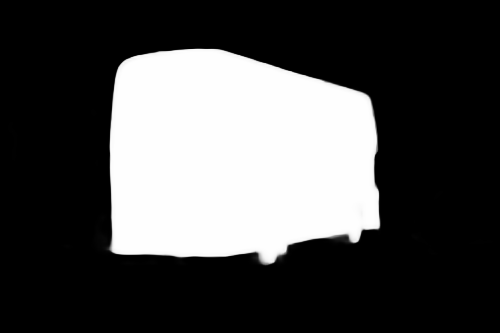}
    \end{minipage}
    \begin{minipage} [t] {0.095\linewidth}
        \centering
        \includegraphics [width=1\linewidth,height=0.75\linewidth] {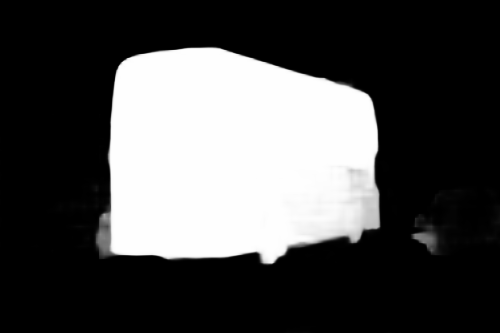}
    \end{minipage}       
    \begin{minipage} [t] {0.095\linewidth}
        \centering
        \includegraphics [width=1\linewidth,height=0.75\linewidth] {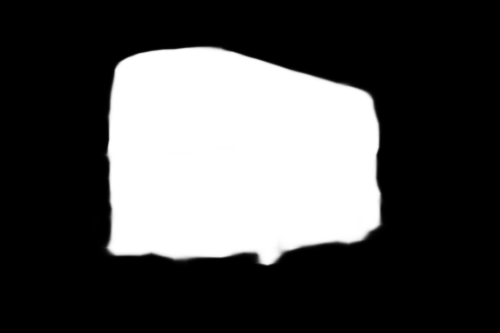}
    \end{minipage}   
    \begin{minipage} [t] {0.095\linewidth}
        \centering
        \includegraphics [width=1\linewidth,height=0.75\linewidth] {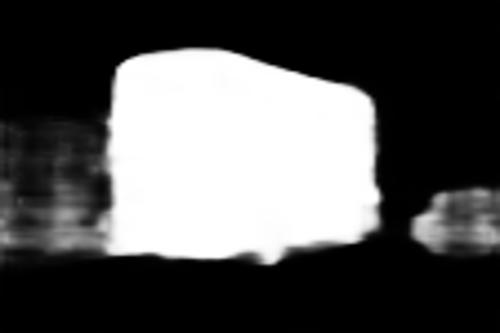}
    \end{minipage} 
    \begin{minipage} [t] {0.095\linewidth}
        \centering
        \includegraphics [width=1\linewidth,height=0.75\linewidth] {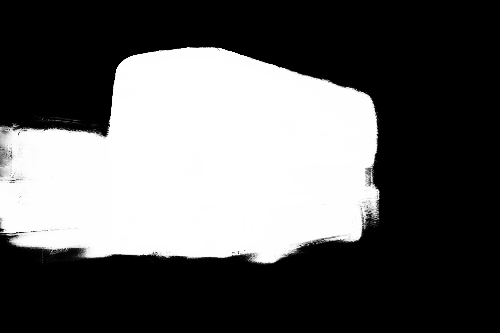}
    \end{minipage}      
    \begin{minipage} [t] {0.095\linewidth}
        \centering
        \includegraphics [width=1\linewidth,height=0.75\linewidth] {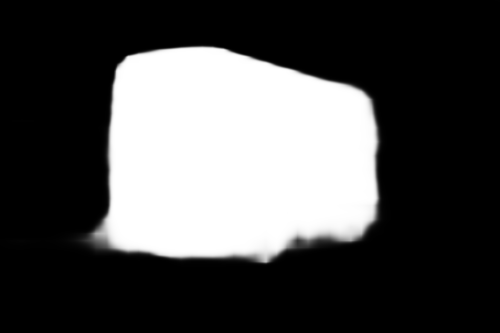}
    \end{minipage}          
    \begin{minipage} [t] {0.095\linewidth}
        \centering
        \includegraphics [width=1\linewidth,height=0.75\linewidth] {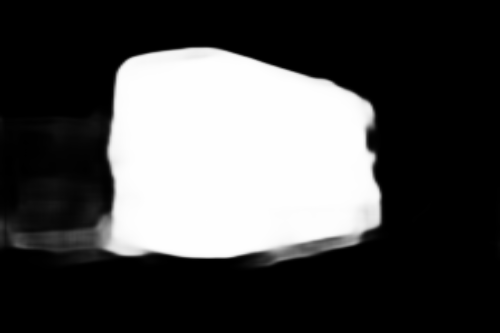}
    \end{minipage}    
    \begin{minipage} [t] {0.095\linewidth}
        \centering
        \includegraphics [width=1\linewidth,height=0.75\linewidth] {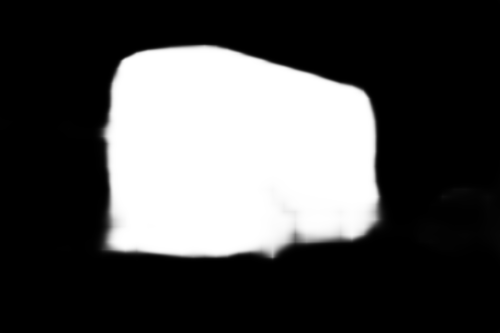}
    \end{minipage}  
    \begin{minipage} [t] {0.095\linewidth}
        \centering
        \includegraphics [width=1\linewidth,height=0.75\linewidth] {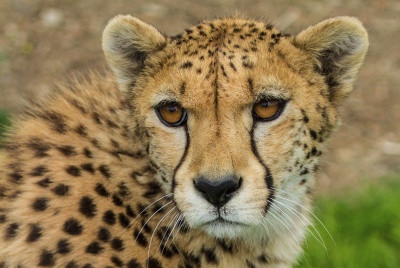}
    \end{minipage}
    \begin{minipage} [t] {0.095\linewidth}
        \centering
        \includegraphics [width=1\linewidth,height=0.75\linewidth] {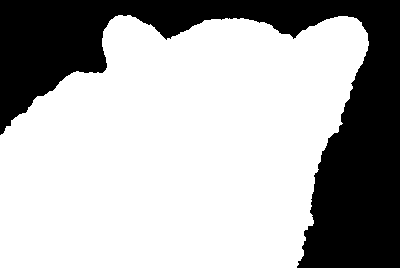}
    \end{minipage}
    \begin{minipage} [t] {0.095\linewidth}
        \centering
        \includegraphics [width=1\linewidth,height=0.75\linewidth] {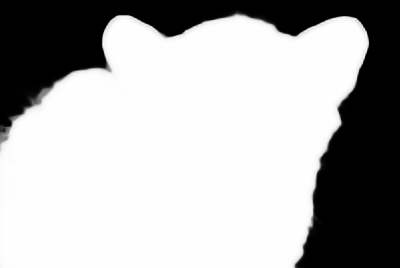}
    \end{minipage}
    \begin{minipage} [t] {0.095\linewidth}
        \centering
        \includegraphics [width=1\linewidth,height=0.75\linewidth] {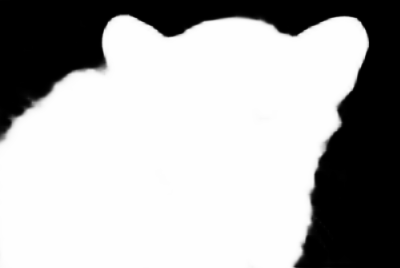}
    \end{minipage}      
    \begin{minipage} [t] {0.095\linewidth}
        \centering
        \includegraphics [width=1\linewidth,height=0.75\linewidth] {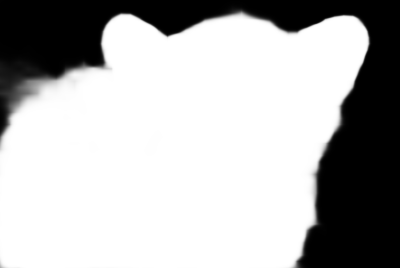}
    \end{minipage}   
    \begin{minipage} [t] {0.095\linewidth}
        \centering
        \includegraphics [width=1\linewidth,height=0.75\linewidth] {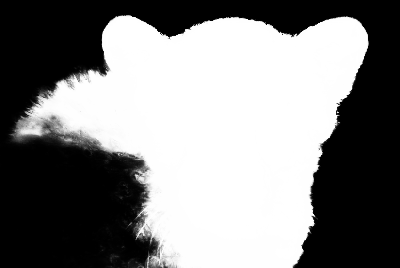}
    \end{minipage} 
    \begin{minipage} [t] {0.095\linewidth}
        \centering
        \includegraphics [width=1\linewidth,height=0.75\linewidth] {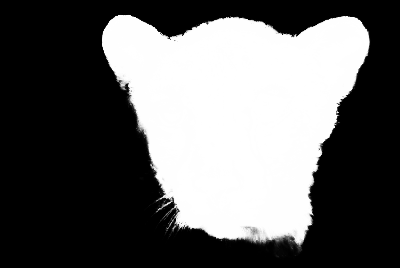}
    \end{minipage}      
    \begin{minipage} [t] {0.095\linewidth}
        \centering
        \includegraphics [width=1\linewidth,height=0.75\linewidth] {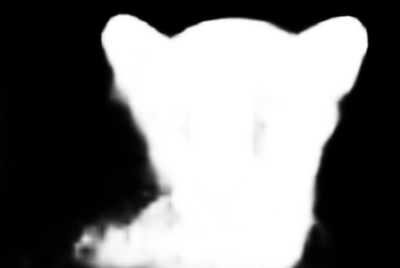}
    \end{minipage}          
    \begin{minipage} [t] {0.095\linewidth}
        \centering
        \includegraphics [width=1\linewidth,height=0.75\linewidth] {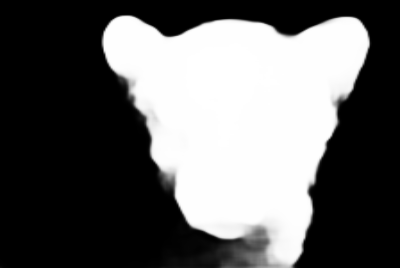}
    \end{minipage}    
    \begin{minipage} [t] {0.095\linewidth}
        \centering
        \includegraphics [width=1\linewidth,height=0.75\linewidth] {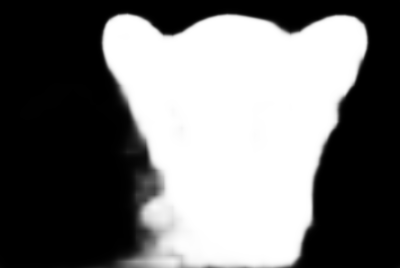}
    \end{minipage}      
    \begin{minipage} [t] {0.095\linewidth}
        \centering
        \includegraphics [width=1\linewidth,height=0.75\linewidth] {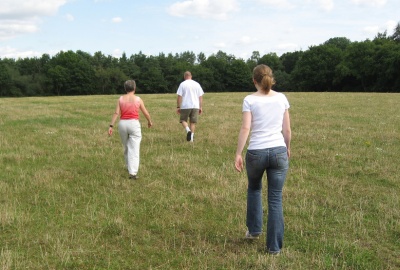}
    \end{minipage}
    \begin{minipage} [t] {0.095\linewidth}
        \centering
        \includegraphics [width=1\linewidth,height=0.75\linewidth] {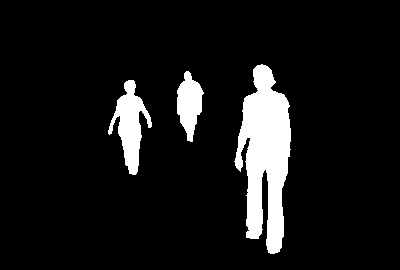}
    \end{minipage}
    \begin{minipage} [t] {0.095\linewidth}
        \centering
        \includegraphics [width=1\linewidth,height=0.75\linewidth] {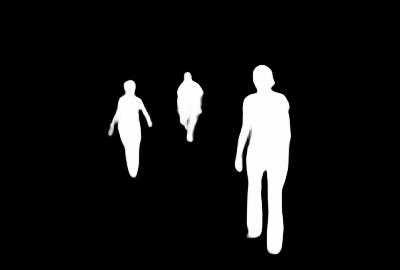}
    \end{minipage}
    \begin{minipage} [t] {0.095\linewidth}
        \centering
        \includegraphics [width=1\linewidth,height=0.75\linewidth] {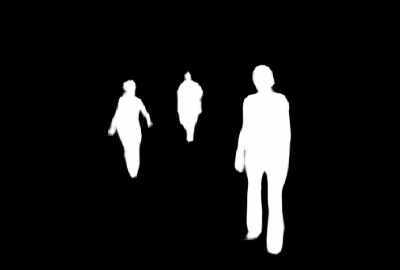}
    \end{minipage}    
    \begin{minipage} [t] {0.095\linewidth}
        \centering
        \includegraphics [width=1\linewidth,height=0.75\linewidth] {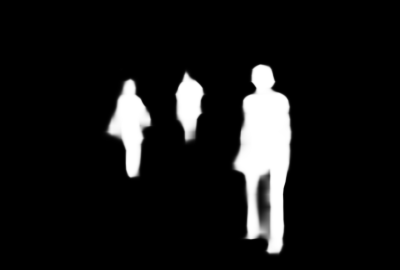}
    \end{minipage}   
    \begin{minipage} [t] {0.095\linewidth}
        \centering
        \includegraphics [width=1\linewidth,height=0.75\linewidth] {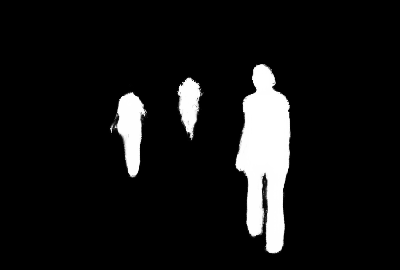}
    \end{minipage} 
    \begin{minipage} [t] {0.095\linewidth}
        \centering
        \includegraphics [width=1\linewidth,height=0.75\linewidth] {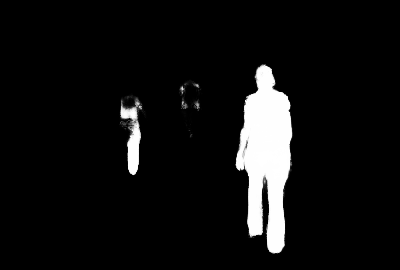}
    \end{minipage}      
    \begin{minipage} [t] {0.095\linewidth}
        \centering
        \includegraphics [width=1\linewidth,height=0.75\linewidth] {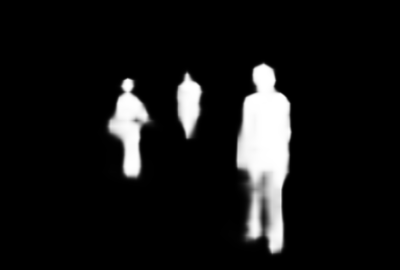}
    \end{minipage}          
    \begin{minipage} [t] {0.095\linewidth}
        \centering
        \includegraphics [width=1\linewidth,height=0.75\linewidth] {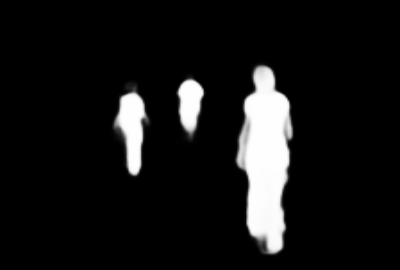}
    \end{minipage}    
    \begin{minipage} [t] {0.095\linewidth}
        \centering
        \includegraphics [width=1\linewidth,height=0.75\linewidth] {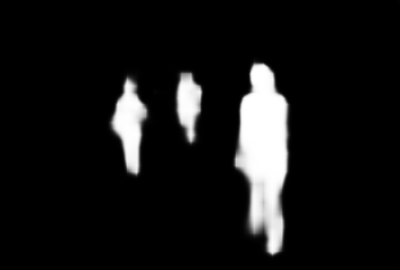}
    \end{minipage}      
    \begin{minipage} [t] {0.095\linewidth}
        \centering
        \includegraphics [width=1\linewidth,height=0.75\linewidth] {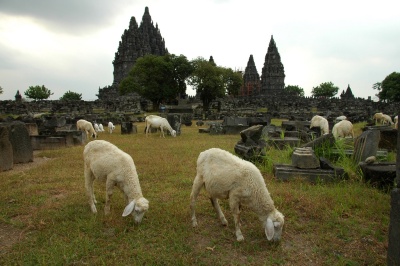}
    \end{minipage}
    \begin{minipage} [t] {0.095\linewidth}
        \centering
        \includegraphics [width=1\linewidth,height=0.75\linewidth] {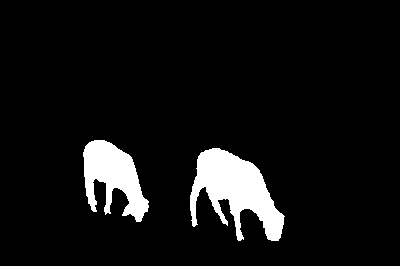}
    \end{minipage}
    \begin{minipage} [t] {0.095\linewidth}
        \centering
        \includegraphics [width=1\linewidth,height=0.75\linewidth] {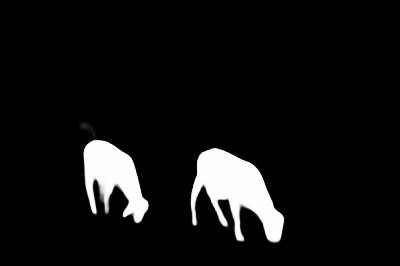}
    \end{minipage}
    \begin{minipage} [t] {0.095\linewidth}
        \centering
        \includegraphics [width=1\linewidth,height=0.75\linewidth] {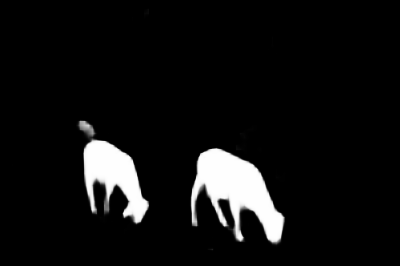}
    \end{minipage}     
    \begin{minipage} [t] {0.095\linewidth}
        \centering
        \includegraphics [width=1\linewidth,height=0.75\linewidth] {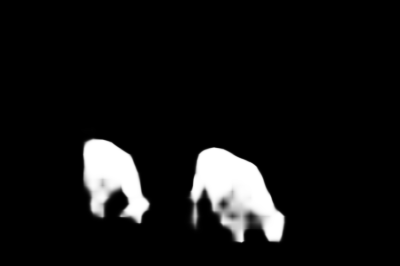}
    \end{minipage}   
    \begin{minipage} [t] {0.095\linewidth}
        \centering
        \includegraphics [width=1\linewidth,height=0.75\linewidth] {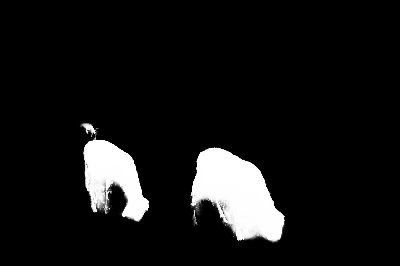}
    \end{minipage} 
    \begin{minipage} [t] {0.095\linewidth}
        \centering
        \includegraphics [width=1\linewidth,height=0.75\linewidth] {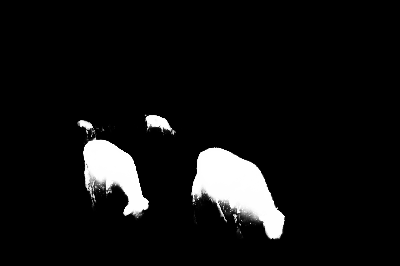}
    \end{minipage}      
    \begin{minipage} [t] {0.095\linewidth}
        \centering
        \includegraphics [width=1\linewidth,height=0.75\linewidth] {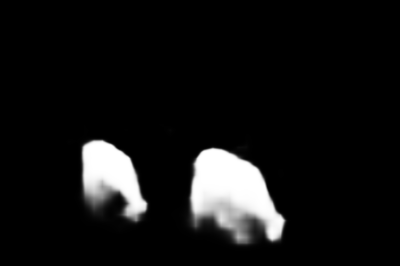}
    \end{minipage}          
    \begin{minipage} [t] {0.095\linewidth}
        \centering
        \includegraphics [width=1\linewidth,height=0.75\linewidth] {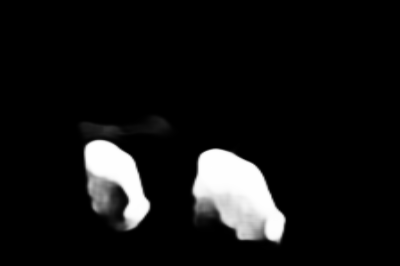}
    \end{minipage}    
    \begin{minipage} [t] {0.095\linewidth}
        \centering
        \includegraphics [width=1\linewidth,height=0.75\linewidth] {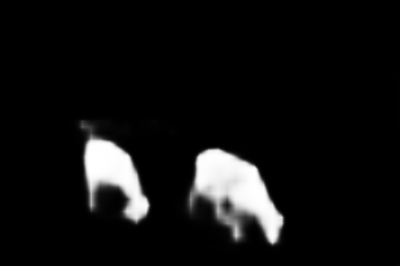}
    \end{minipage}     
    \begin{minipage} [t] {0.095\linewidth}
        \centering
        \includegraphics [width=1\linewidth,height=0.75\linewidth] {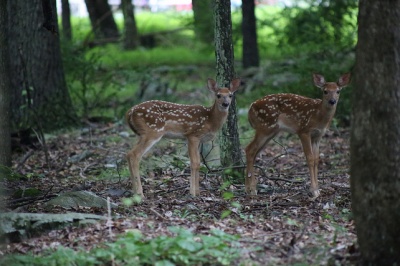}
        {(a)}
    \end{minipage}
    \begin{minipage} [t] {0.095\linewidth}
        \centering
        \includegraphics [width=1\linewidth,height=0.75\linewidth] {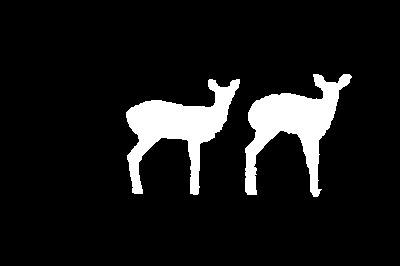}
        {(b)}
    \end{minipage}
    \begin{minipage} [t] {0.095\linewidth}
        \centering
        \includegraphics [width=1\linewidth,height=0.75\linewidth] {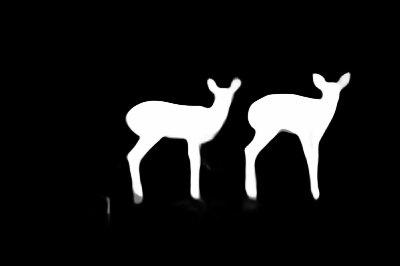}
        {(c)}
    \end{minipage}
    \begin{minipage} [t] {0.095\linewidth}
        \centering
        \includegraphics [width=1\linewidth,height=0.75\linewidth] {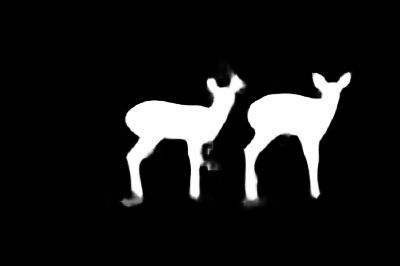}
        {(d)}
    \end{minipage} 
    \begin{minipage} [t] {0.095\linewidth}
        \centering
        \includegraphics [width=1\linewidth,height=0.75\linewidth] {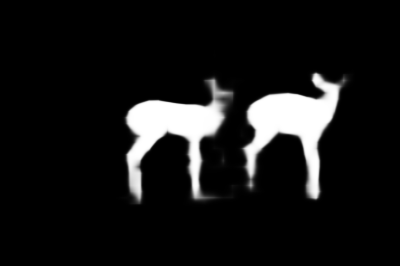}
        {(e)}
    \end{minipage}   
    \begin{minipage} [t] {0.095\linewidth}
        \centering
        \includegraphics [width=1\linewidth,height=0.75\linewidth] {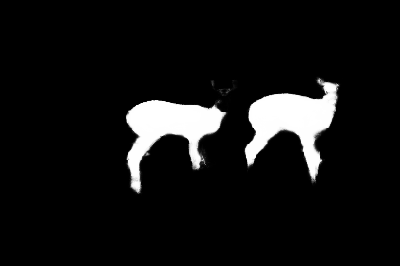}
        {(f)}
    \end{minipage} 
    \begin{minipage} [t] {0.095\linewidth}
        \centering
        \includegraphics [width=1\linewidth,height=0.75\linewidth] {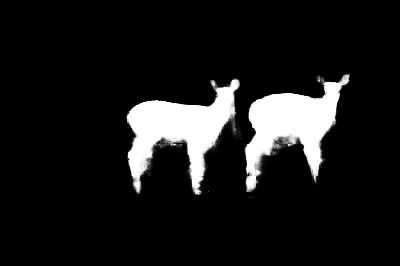}
        {(g)}
    \end{minipage}      
    \begin{minipage} [t] {0.095\linewidth}
        \centering
        \includegraphics [width=1\linewidth,height=0.75\linewidth] {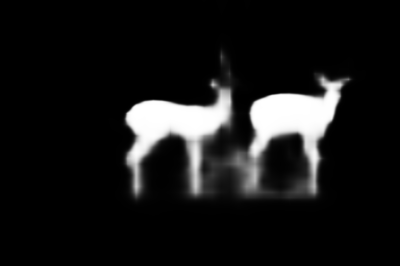}
        {(h)}
    \end{minipage}          
    \begin{minipage} [t] {0.095\linewidth}
        \centering
        \includegraphics [width=1\linewidth,height=0.75\linewidth] {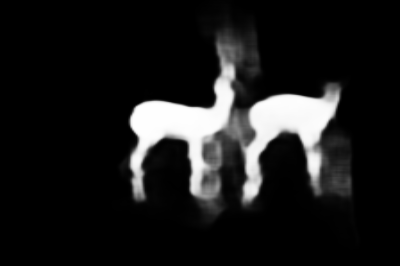}
        {(i)}
    \end{minipage}    
    \begin{minipage} [t] {0.095\linewidth}
        \centering
        \includegraphics [width=1\linewidth,height=0.75\linewidth] {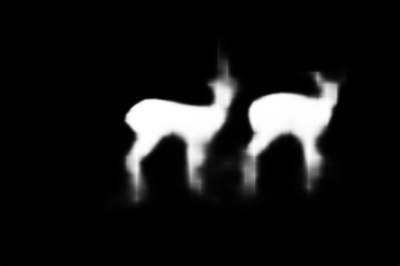}
        {(j)}
    \end{minipage}     
    \vspace{-0.2cm}
    \caption{Qualitative comparison with previous state-of-the-art methods. (a)  {Input image}. (b)  {Ground truth}. (c)  {Baseline+BRL+RRM}. (d) { Baseline+BRL}. 
    (e)  {CPD-R} \cite{cascadesodcvpr2019}. (f)  {PiCANet-R} \cite{liu2018PicaNet}. (g)  {R3-Net} \cite{r3net}. (h) {DGRL} \cite{exp6}. (i)  {C2S-Net} \cite{contour}.
    (j)  {SRM} \cite{srm}. Results of R3-Net are refined by CRF \cite{CRF}.}
    \label{fig:more_sample}
\end{figure}

\vspace{-0.1cm}
More visual comparisons are shown in Figure \ref{fig:more_sample}.

\subsection{Links for Datasets}
\label{sec:links}
\subsubsection{Salient Object Detection}
\textbf{HKU-IS} \cite{HKU-IS}: https://i.cs.hku.hk/\textasciitilde gbli/deep\_saliency.html\\
\textbf{ECSSD} \cite{ECSSD}: http://www.cse.cuhk.edu.hk/leojia/projects/hsaliency/dataset.html \\
\textbf{DUTS} \cite{DUTS}: http://saliencydetection.net/duts/ \\
\textbf{DUT-OMRON} \cite{OMRON}: http://saliencydetection.net/dut-omron/ \\
\textbf{PASCAL-S} \cite{PASCAL-S}: http://www.cbi.gatech.edu/salobj/ \\
\textbf{MSRA-B} \cite{msra-b} \& \textbf{MSRA-10K} \cite{contrast1}: https://mmcheng.net/zh/msra10k/

\vspace{-0.2cm}
\subsubsection{Shadow Detection}
\textbf{UCF} \cite{ucf}: http://aqua.cs.uiuc.edu/site/projects/shadow.html \\
\textbf{SBU} \cite{sbu1,sbu2}: https://www3.cs.stonybrook.edu/\textasciitilde cvl/dataset.html \\
\textbf{ISTD} \cite{istd}: https://drive.google.com/file/d/1I0qw-65KBA6np8vIZzO6oeiOvcDBttAY/view 

\vspace{-0.2cm}
\subsubsection{Portrait Segmentation}
\textbf{Flickr Portrait Dataset\cite{portrait}}: http://xiaoyongshen.me/webpage\_portrait/index.html

\clearpage
\medskip
{\small
    \bibliographystyle{ieee}
    \bibliography{bib}
}

\end{document}